\documentclass[11pt]{article}

\usepackage[T1]{fontenc}
\usepackage{amsmath}
\usepackage{amsfonts}
\usepackage{amscd}
\usepackage{amssymb}
\usepackage{amsthm}
\usepackage{bm}
\usepackage{thm-restate}

\usepackage{graphicx}
\usepackage{fontawesome5}

\usepackage{enumerate}
\usepackage{enumitem}

\usepackage{epsfig}

\usepackage[usenames,dvipsnames]{xcolor}

\usepackage{url}

\usepackage{booktabs,multirow,multicol}

\usepackage{natbib}
\usepackage[parfill]{parskip}

\usepackage[colorlinks,linktoc=all]{hyperref}
\hypersetup{citecolor=MidnightBlue}
\hypersetup{urlcolor=MidnightBlue}
\hypersetup{linkcolor=Black}

\usepackage[nameinlink]{cleveref}
\creflabelformat{equation}{#1#2#3}
\crefname{equation}{eq.}{eqs.}
\Crefname{equation}{Eq.}{Eqs.}
\Crefname{section}{Section}{Section}

\DeclareRobustCommand{\parhead}[1]{\textbf{#1}~}

\numberwithin{equation}{section}

\usepackage{caption}
\usepackage{subcaption}
\newtheorem{definition}{Definition}

\newtheorem{remark}{Remark}

\usepackage{tcolorbox}

\usepackage{amsmath,amsfonts,bm}

\def\1{\bm{1}}

\DeclareMathAlphabet{\mathsfit}{\encodingdefault}{\sfdefault}{m}{sl}
\SetMathAlphabet{\mathsfit}{bold}{\encodingdefault}{\sfdefault}{bx}{n}

\newcommand{\E}{\mathbb{E}}

\newcommand{\R}{\mathbb{R}}

\usepackage[algoruled,ruled,vlined]{algorithm2e}
\usepackage{listings}
\usepackage{fancyvrb}
\fvset{fontsize=\normalsize}

\makeatletter
\renewcommand{\SetKwInOut}[2]{\sbox\algocf@inoutbox{\KwSty{#2}\algocf@typo:}\expandafter\ifx\csname InOutSizeDefined\endcsname\relax \newcommand\InOutSizeDefined{}\setlength{\inoutsize}{\wd\algocf@inoutbox}\sbox\algocf@inoutbox{\parbox[t]{\inoutsize}{\KwSty{#2}\algocf@typo:\hfill}~}\setlength{\inoutindent}{\wd\algocf@inoutbox}\else \ifdim\wd\algocf@inoutbox>\inoutsize \setlength{\inoutsize}{\wd\algocf@inoutbox}\sbox\algocf@inoutbox{\parbox[t]{\inoutsize}{\KwSty{#2}\algocf@typo:\hfill}~}\setlength{\inoutindent}{\wd\algocf@inoutbox}\fi \fi \algocf@newcommand{#1}[1]{\ifthenelse{\boolean{algocf@inoutnumbered}}{\relax}{\everypar={\relax}}{\let\\\algocf@newinout\hangindent=\inoutindent\hangafter=1\parbox[t]{\inoutsize}{\KwSty{#2}\algocf@typo:\hfill}~##1\par}\algocf@linesnumbered }}\makeatother

\SetKwInOut{KwInput}{input}
\SetKwInOut{KwOutput}{output} 

\usepackage{booktabs}
\usepackage{commath}
\usepackage{enumitem}
\usepackage{pgf, tikz}
\usetikzlibrary{arrows, automata}

\usepackage[export]{adjustbox}

\DeclareFontFamily{U}{mathx}{\hyphenchar\font45}
\DeclareFontShape{U}{mathx}{m}{n}{
      <5> <6> <7> <8> <9> <10> gen * mathx
      <10.95> mathx10 <12> <14.4> <17.28> <20.74> <24.88> mathx12
      }{}
\DeclareSymbolFont{mathx}{U}{mathx}{m}{n}
\DeclareMathSymbol{\intop}  {1}{mathx}{"B3}

\DeclareFontFamily{U}{mathx}{\hyphenchar\font45}
\DeclareFontShape{U}{mathx}{m}{n}{
      <5> <6> <7> <8> <9> <10>
      <10.95> <12> <14.4> <17.28> <20.74> <24.88>
      mathx10
      }{}
\DeclareSymbolFont{mathx}{U}{mathx}{m}{n}
\DeclareFontSubstitution{U}{mathx}{m}{n}
\DeclareMathAccent{\widecheck}{0}{mathx}{"71}
\DeclareMathAccent{\wideparen}{0}{mathx}{"75}

\newcommand{\wh}{\widehat}

\let\temp\phi
\let\phi\varphi
\let\varphi\temp

            \newcommand{\given}{\,|\,}

\newcommand{\gr}{\mathsf{G}}
\newcommand{\ver}{\mathsf{V}}
\newcommand{\edg}{\mathsf{E}}
\DeclareMathOperator{\pa}{pa}

\newcommand{\probs}{\mathcal{P}}

\newcommand{\funcs}{\mathcal{F}}

\newcommand{\mixcomp}[2]{C}
\newcommand{\mixwgt}[2]{\pi}

\newcommand{\bx}{\bm{x}}
\newcommand{\bz}{\bm{z}}

\newcommand{\DAG}{\text{DAG}}

\newcommand{\latsem}{\bm{A}}
\newcommand{\latsemwt}{a}
\newcommand{\grB}{\mathsf{B}}

\newcommand{\latgr}{\gr}
\newcommand{\bn}{\grB}

\newcommand{\xdim}{D}
\newcommand{\zdim}{K}

\newcommand{\zpr}{p_{\bz}}
\newcommand{\zinv}{p_{\bz}^{(e)}}
\newcommand{\zprs}{\probs}

\newcommand{\fcns}{\mathcal{F}}

\usepackage{tikz}
\usepackage{tkz-graph}
\usetikzlibrary{tikzmark,decorations.pathreplacing,calligraphy,arrows.meta, positioning,fit,calc}

\usepackage{soul}

\newcommand{\disablecomments}{
    \newcommand{\change}[1]{\textcolor{blue}{##1}}
    \newcommand{\gemma}[1]{}
    \newcommand{\bryon}[1]{}
    \newcommand{\bryonnew}[1]{}
    \newcommand{\hilite}[1]{\textcolor{red}{##1}}
}

\disablecomments

\newcommand{\blind}{1}

\begin{document}

\newcommand{\thetitle}{Towards Interpretable Deep Generative Models via Causal Representation Learning}

\if1\blind
{
  \title{\bf \thetitle}
  \author{Gemma Moran\\
    Department of Statistics, Rutgers University\\
    and \\
    Bryon Aragam\\
    Booth School of Business, University of Chicago}
  \maketitle
} \fi

\if0\blind
{
  \bigskip
  \bigskip
  \bigskip
  \begin{center}
    {\LARGE\bf \thetitle}
\end{center}
  \medskip
} \fi

\bigskip
\begin{abstract}

Recent developments in generative artificial intelligence (AI) rely on machine learning techniques such as deep learning and generative modeling to achieve state-of-the-art performance across wide-ranging domains.  These methods' surprising performance is due in part to their ability to learn implicit ``representations'' of complex, multi-modal data. Unfortunately, deep neural networks are notoriously black boxes that obscure these representations, making them difficult to interpret or analyze. To resolve these difficulties, one approach is to build new interpretable neural network models from the ground up. This is the goal of the emerging field of causal representation learning (CRL) that uses causality as a vector for building flexible, interpretable, and transferable generative AI. CRL can be seen as a synthesis of three intrinsically statistical ideas: (i) latent variable models such as factor analysis; (ii) causal graphical models with latent variables; and (iii) nonparametric statistics and deep learning. This paper introduces CRL from a statistical perspective, focusing on connections to classical models as well as statistical and causal identifiability results. We also highlights key application areas, implementation strategies, and open statistical questions.  
 \end{abstract}

\vfill

\newpage

\section{Introduction}

Generative AI has achieved state-of-the-art performance on a variety of tasks across numerous domains; for example, predicting protein folding \citep{jumper2021highly}, generating new candidate molecules for drug discovery \citep{merchant2023scaling}, text-to-image generation \citep{betker2023improving} and test taking \citep{openai2024gpt4technicalreport}. 
This performance has been driven by machine learning (ML) techniques such as deep learning and generative modeling.

These methods' surprising performance is due in part to their ability to learn implicit ``representations'' of complex, multi-modal data. 
A representation is simply a latent numerical summary of data that encodes semantically meaningful features, and can be interpreted analogously to latent factors in factor analysis. Indeed, this interpretation is explicit in earlier work on probabilistic machine learning \citep{roweis1999unifying,lawrence2005gplvm,goodfellow2012large}.
This early work focused on latent variable models such as Gaussian process latent variable models (GPLVMs), mixtures, and hierarchical Bayesian models; however, the past decade has witnessed a shift towards deep learning, where nonlinearities are captured with deep neural networks of essentially arbitrary complexity \citep{rezende2014vae,kingma2013auto,ranganath2014bbvi}.

Unfortunately, as is well-known, neural networks are ``black boxes'' that obscure such data representations, 
making them difficult to even define concretely, let alone interpret or analyze \citep{rudin2019stop}. Without understanding how ML algorithms are learning and manipulating data representations, it is difficult to trust that these algorithms are performing as they should. Moreover, understanding such algorithms is crucial to gaining scientific insight into the data problem at hand; performance in prediction and generation is not enough. 

To resolve these difficulties in understanding ML algorithms, there are two main directions in which to proceed. The first direction is to develop tools to interpret and analyze pre-trained models (often called ``post-hoc explainability''). Such post-hoc tools and analyses have already provided insight into image and language models \citep{selvaraju2017grad,templeton2024scaling}. A problem with such post-hoc explainability methods, however, is that they may not accurately reflect what the original model computes  \citep{rudin2019stop}; moreover, many explainability methods are provably no better than random guessing at inferring model behavior \citep{bilodeau2024impossibility}. As an alternative, the second direction is to build new models which have more transparent and interpretable internal mechanisms. By building interpretable models ``from the ground up'', the goal is to obtain more confidence in and understanding of their conclusions.

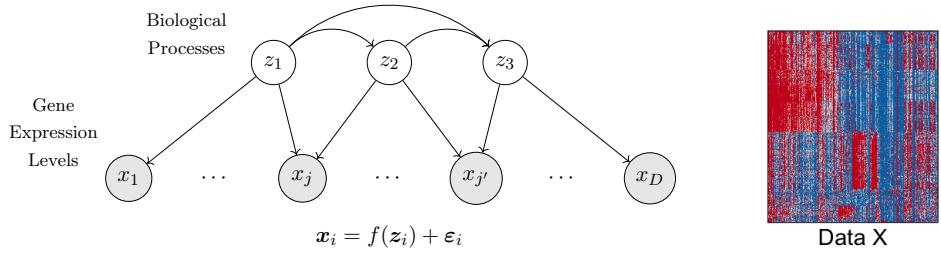
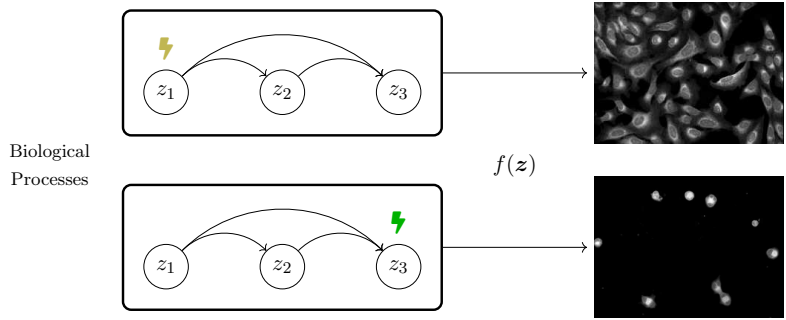
\begin{figure}[ht!]
\centering
\begin{subfigure}[b]{\textwidth}
\centering
\resizebox{0.6\textwidth}{!}{  
\begin{tikzpicture}[
    vertex/.style={circle, draw, minimum size=20pt},
    zlabel/.style={vertex, fill=white},
    xlabel/.style={vertex, fill=gray!20}
]

\foreach \i in {1,2,3} {
    \node[zlabel] (z\i) at (2*\i-4,2) {$z_{\i}$};
}

\node[align=center] at (-3.5,2.5) {\footnotesize Biological \\ \footnotesize Processes};

\node[align=center] at (-5.8,0.8) {\footnotesize Gene \\ \footnotesize Expression \\\footnotesize Levels};

\node[xlabel] (x1) at (-4.5, 0) {$x_1$};
\node[] at (-3, 0) {$\cdots$};
\node[xlabel] (x3) at (-1.5, 0) {$x_j$};
\node[] at (0,0) {$\cdots$};
\node[xlabel] (x5) at (1.5, 0) {$x_{j'}$};
\node[] at (3, 0) {$\cdots$};
\node[xlabel] (x7) at (4.5, 0) {$x_\xdim$};

\node[] (f) at (0,-1) {$\bx_{i}=f(\bz_i)+\bm{\varepsilon}_i$};

\draw[->] (z1) -- (x1);
\draw[->] (z1) -- (x3);
\draw[->] (z2) -- (x3);
\draw[->] (z2) -- (x5);
\draw[->] (z3) -- (x5);
\draw[->] (z3) -- (x7);

\draw[->, bend left=45] (z1) to (z2);
\draw[->, bend left=45] (z1) to (z3);
\draw[->, bend left=45] (z2) to (z3);

\end{tikzpicture} 
 }
\hspace{0.75cm}
\includegraphics[width=0.16\textwidth]{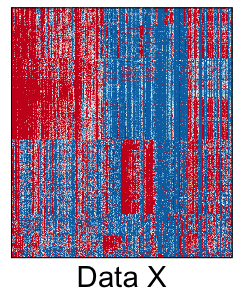}
\caption{{\bf Example 1: Gene expression data.} Left: Graphical model  where the expression of  gene $x_j$ is drawn from  latent low-dimensional factors $z_k$ with dependence among the latent $z_k$. Right: Matrix of gene expression data from \citet{zeisel2015cell} (subset of 500 cells by 300 genes).}
\label{fig:intro-example-1}
\end{subfigure}

\begin{subfigure}[b]{\textwidth}
\centering
\resizebox{0.7\textwidth}{!}{

\begin{tikzpicture}[
    vertex/.style={circle, draw, minimum size=12pt},
    zlabel/.style={vertex, fill=white}
]
\foreach \i in {1,2,3} {
    \node[zlabel] (z\i) at (2*\i-4,2) {$z_{\i}$};
}
\node (bolt) at (-2,2.75) {\textcolor{yellow!70!black}{\faIcon{bolt}}};

\path (z1) -- (z2) coordinate[midway, yshift=30pt] (m12);
\path (z1) -- (z3) coordinate[midway, yshift=30pt] (m13);
\path (z2) -- (z3) coordinate[midway, yshift=30pt] (m23);

\draw[->, bend left=45] (z1) to (z2);
\draw[->, bend left=45] (z1) to (z3);
\draw[->, bend left=45] (z2) to (z3);
\node[draw, rounded corners, very thick, fit=(z1) (z2) (z3) (m12) (m13) (m23), inner sep=10pt] (box) {};

\node (image1) at (box -| 7,0) {\includegraphics[width=3.25cm]{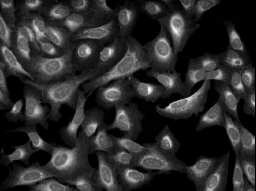}};

\draw[->] (box.east) -- node[above, midway] {} (image1);

\foreach \i in {1,2,3} {
    \node[zlabel] (z\i b) at (2*\i-4,-1) {$z_{\i}$};
}
\node (bolt) at (2,-0.25) {\textcolor{green!70!black}{\faIcon{bolt}}};
\path (z1b) -- (z2b) coordinate[midway, yshift=30pt] (m12b);
\path (z1b) -- (z3b) coordinate[midway, yshift=30pt] (m13b);
\path (z2b) -- (z3b) coordinate[midway, yshift=30pt] (m23b);

\node[align=center] at (-4,0.75) {\footnotesize Biological \\ \footnotesize Processes};

\node[align=center] at (4,0.75) {$f(\bm{z})$};

\draw[->, bend left=45] (z1b) to (z2b);
\draw[->, bend left=45] (z1b) to (z3b);
\draw[->, bend left=45] (z2b) to (z3b);
\node[draw, rounded corners, very thick, fit=(z1b) (z2b) (z3b) (m12b) (m13b) (m23b), inner sep=10pt] (boxb) {};
\node (image2) at (boxb -| 7,0) {\includegraphics[width=3.25cm]{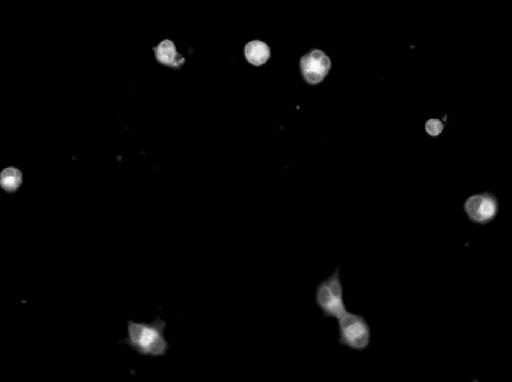}};

\draw[->] (boxb.east) -- node[above, midway] {} (image2);

\end{tikzpicture}
 
 }
\caption{{\bf Example 2: Cell image data.} A generative model for images of perturbed cells. In each image, different compounds have been added to the cells, represented as a perturbation of the latent features of the cell. Images from \citet{bray2017image} (top: CIL45758; bottom: CIL45766). 
}  
\label{fig:intro-example-2}
\end{subfigure}

\caption{Examples of generative models. The causal representation learning problem is to recover the latent factors along with the dependence structure, both of which are unknown, from the observed features. The nonlinear mapping $f$ can be estimated using a deep neural network with many more hidden units than the number of latent factors.
}\label{fig:intro}
\end{figure}
One first step to building more interpretable models is to  consider a generative model where the latent representations are explicitly parameterized, as in  \Cref{fig:intro} (a more formal development will be deferred until later sections). Here, the training data is given by a high-dimensional random vector $\bm{x}_{i}$, and we seek a latent ``representation'' of $\bm{x}_{i}$ via some hidden variables $\bm{z}_{i}$. We assume that $\bm{x}_{i}$ is generated from $\bm{z}_{i}$ via an (unknown) function $f$. In this setting, $\bm{z}_{i}$ encodes the concept of a ``learned representation''. As an example, $\bm{x}_{i}$ may be a vector of gene expression values, and the $\bm{z}_{i}$ may represent different biological processes. In this generative model, switching ``on'' a dimension of $\bm{z}_{i}$ will then impact the expression levels of genes associated with that biological process. As another example, $\bm{x}_{i}$ may be pixels of images and $\bm{z}_{i}$ may correspond to image features such as content and style. In an ideal interpretable model, changing the dimension of $\bm{z}_{i}$ corresponding to ``style'' will generate images with different stylistic choices while keeping the content of the image fixed.

Many modern deep generative models are explicitly of this form, including autoencoders, diffusions and flows. The study of such generative models, however, has focused mainly on the problem of \emph{generation}; equivalently, sampling or density estimation. That is, the goal is simply to produce realistic samples from complex distributions over text, images, audio, and/or video. Crucially, it is possible to sample from a distribution as a black-box without any conceptual understanding of underlying mechanisms or processes. As a result, the problem of \emph{understanding} generative models, especially in terms of estimation of the parameters $(\bm{z}_{i},f)$, has been less well studied. 
This distinction between density and parameter estimation is not unique to generative models, and parallels the ``two cultures'' of statistical modeling, namely prediction-focused and inference-focused modeling \citep{breiman2001statistical}; see \Cref{sec:crl} for more discussion.

To understand and interpret generative models as in \Cref{fig:intro}, parameter estimation is crucial. A key technical challenge in this endeavor is that generative models with different parameters can still produce identical density estimates, despite their internal mechanisms being completely different. Phrased more formally, the parametrization of generative models given in \Cref{fig:intro} is \emph{non-identifiable}. Moreover, this non-identifability makes it difficult to certify the behavior of a generative model using post-hoc explainability techniques. To render models identifiable, we need to either a) restrict the form of the model, or b) gain access to richer data modalities arising from different environments. 

In classical analogues of representation learning, such as factor models and independent component analysis, identifiability and interpretable parameter estimation are well-studied. (Later, in \Cref{sec:linear}, we outline the connections between generative models and linear factor models in particular.)  However, such classical approaches with restrictive parametric models are often insufficient to model complex unstructured and multimodal datasets that proliferate in modern applications of AI.
In these settings, more flexible models that avoid these restrictions are needed. Further, there is increased demand for models to transfer well to potentially unspecified downstream tasks (as opposed to specializing to specific, well-specified tasks \emph{a priori}, which is often the case for classical models).

Motivated by these goals of flexibility, interpretability, and transferability, the field of \emph{causal representation learning} (CRL) has emerged to tackle these issues by bringing the flexibility of deep learning together with the interpretability and transferability of causality.  It is widely believed that modeling  \emph{causal} relationships results in stable, robust, and invariant ML models that are easier to interpret \citep{peters2016causal,buhlmann2020invariance}. Further, the value of causal models for stable generalization and transfer to unseen tasks is widely recognized in the ML community. Indeed, the position paper by \citet{scholkopf2021towards} argued that causality is needed for meaningful representation learning; the current article targets recent progress in this direction.

As much of this causal representation learning literature has evolved independently in ML conferences such as NeurIPS, ICML, and ICLR, this article is intended to provide a general overview of this intrinsically statistical problem to the broader statistical community. Our goal is not to provide a comprehensive survey of papers, of which there are far too many, but to explain and contextualize this problem within the framework of statistics.  

\noindent
\parhead{Outline}
In order to keep the presentation self-contained and provide context for statisticians, we begin by introducing two motivating examples in \Cref{sec:motiv} and then review classical results on linear factor analysis in \Cref{sec:linear}, including a discussion of its limitations for generative AI. This will motivate consideration of causal models, which are reviewed in \Cref{sec:causal}. Then \Cref{sec:crl} formally introduces the problem of CRL, and \Cref{sec:nonlinear} discusses existing identifiability results in nonlinear CRL models, which introduce a host of technical complications. \Cref{sec:practical} discusses practical challenges in estimation. Finally, we conclude in \Cref{sec:conc} with a discussion of recent work and open problems.

\section{Motivating examples}
\label{sec:motiv}

The idea of identifying latent factors with causal implications is at the heart of scientific discovery. To help ground our discussion, the 
following two examples serve to motivate some of the key ideas behind CRL.

\noindent
\parhead{Motivating example 1: Genomics.} 
Perturb-Seq \citep{dixit2016perturb} combines single-cell RNA sequencing with CRISPR-based perturbations of specific genes. The data consists of an observational dataset $\bm{X}\in\mathbb{R}^{N\times \xdim}$ of $N$ cells and $\xdim$ genes, and interventional datasets $\bm{X}^{(e)}\in \mathbb{R}^{N_e \times \xdim}$, $e\in\mathcal{E}$. In each intervention $e$, a particular gene  is targeted by CRISPR perturbation. Perturbation types include gene knockouts \citep[deletion or disruption of target genes via CRISPR-Cas 9, ][]{cong2013multiplex}, gene knockdowns \citep[reduction in gene expression via CRISPRi, ][]{QI20131173} or gene activation \citep[CRISPRa to enhance gene expression, ][]{gilbert2014genome}. 
See \Cref{fig:intro-example-1}.

These interventional data provide an opportunity to better understand gene regulatory networks. One approach to analyzing this data is to analyze the gene expression levels directly, using linear regression on the perturbation indicator or causal discovery algorithms. However, such approaches do not take advantage of the latent low-dimensional structure in the data. A potentially more efficient approach to causal discovery is to reduce the dimension of the data and consider the causal graph over the latent space.  For example, if a set of genes exhibit coordinated expression, representing a gene program, we can consider these genes as a single causal variable and seek to interpret its biological function. See \citet{zhang2023identifiability} for a real data example with Perturb-Seq data in the context of CRL.

\noindent
\parhead{Motivating example 2: Images.} 
Images are difficult to manage at the pixel-level: For example, individual pixels do not carry causal content (e.g. intervening on a pixel is generally not meaningful in applications). Generative models can be used to learn latent representations that better capture factors of variation in images (e.g. colour, shape, size), and in particular, latent factors that capture causal dependencies. 
Consider image data $\bm{X}\in\mathbb{R}^{N\times \xdim}$ of $N$ images and $\xdim$ pixels, and interventional datasets $\bm{X}^{(e)}\in \mathbb{R}^{N_e \times \xdim}$, $e\in\mathcal{E}$. 
In cell microscopy images, for example, each intervention $e$ may correspond to CRISPR knockouts \citep{jain2024automated} or drug administration \citep{chandrasekaran2021image}. 
More broadly, in general imaging and vision applications, interventions $e$ can correspond to different forms of data augmentation, e.g. images can be rotated, translated, rescaled, colour corrected, or texturized. 
In these applications, the effects of data augmentation or a CRISPR knockout cannot be detected at the individual pixel level, and instead these effects act on latent factors to be inferred from the data. 
See \citet{von2021self,lippe2022citris,brehmer2022weakly} for imaging applications in the context of CRL.

In both of these examples, learning latent causal factors can aid in interpretability, computational efficiency, and potentially improve power to detect effects via data pooling, but there are a number of challenges. Are these latent abstractions identifiable and interpretable? When is it possible to obtain the causal graph of the latent variables? These are the goals of causal representation learning. 
Before introducing this problem formally, we first review some preliminaries on factor models (Section~\ref{sec:linear}) and causality (Section~\ref{sec:causal}) from the perspective of CRL.

\section{Linear Representation Learning}\label{sec:linear}

Given the close connection between representation learning and factor analysis, we begin by briefly reviewing classical linear factor models.
Our purpose here is not to be exhaustive, but rather to provide context for later sections.
In the linear factor analysis model, the observed data is a vector of features $\bm{x}_i\in\R^\xdim$ for each sample $i\in[N]:=\{1,\ldots,N\}$. Each observation $\bm{x}_i$ is assumed to be generated from a lower dimensional vector of latent factors $\bm{z}_i \in \R^\zdim$ ($\xdim \gg \zdim$): 
\begin{align}
\bm{x}_i &= \bm{B}\bm{z}_i + \bm{\varepsilon}_i,\quad \bm{\varepsilon}_i \sim \mathcal{N}_D(\bm{0}, \bm{\Sigma}), 
\quad \bm{z}_i \sim \zpr(\bm{z}_i),
\label{eq:factor-analysis}
\end{align}
where $\bm{B}\in\mathbb{R}^{\xdim \times \zdim}$ is the factor loadings matrix and $\bm{\Sigma}=\text{diag}(\sigma_1^2,\dots, \sigma_\xdim^2)$. This is a special case of the general model in \Cref{fig:intro} where the transformation $f$ is linear.

\begin{figure}[t]
    \centering
    \begin{subfigure}[t]{0.65\textwidth}
    \centering
    \includegraphics[width=\textwidth]{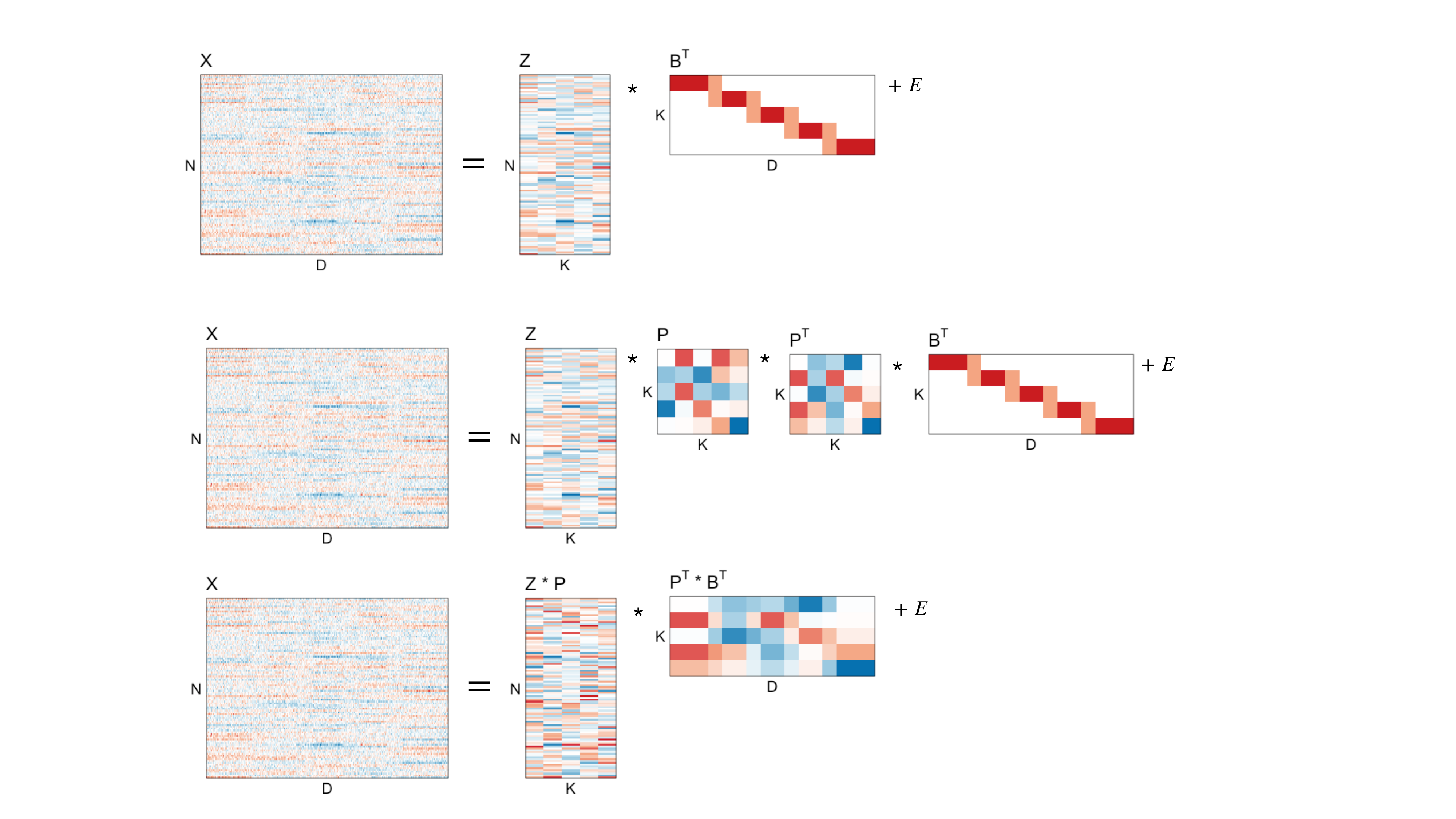}
    \caption{} 
    \label{fig:linear-model}
    \end{subfigure}
    \hspace{0.1cm}
    \begin{subfigure}[t]{0.25\textwidth}
    \centering 
    \includegraphics[width=0.8\textwidth]{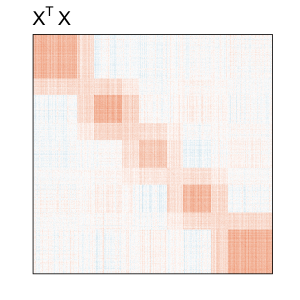}
    \caption{} 
    \label{fig:correlation}
    \end{subfigure}
    \caption{A simulated dataset from the linear factor analysis model with $N=100$ samples, $\xdim=1956$ features, and $\zdim=5$ factors. (a) The data $\bm{X}$, latent factors $\bm{Z}$ and loadings $\bm{B}$; (b) The empirical covariance matrix $\bm{X}^\top\bm{X}/n$.}
\end{figure}

A common goal of linear factor analysis is to find ``scientifically meaningful'' groups of features. In the linear factor analysis model, ``scientifically meaningful'' is operationalized as a group of features which covary. These covarying groups can be inferred from the loadings matrix, $\bm{B}$. Specifically, the $k$th group of features corresponds to the non-zero elements of $\bm{B}_{\cdot, k}$, the $k$th column of the loadings matrix (\Cref{fig:linear-model}).  

\noindent
\parhead{Identifiability.}  The goal of finding meaningful factors is hindered by the well-known issue of identifiability in \eqref{eq:factor-analysis}. Specifically, the linear factor analysis model is not identifiable when $\zpr= \mathcal{N}_K(\bm{0}, \bm{I})$. 
To see this, suppose $\bm{z}_i$ and $\bm{B}$ are the true parameters that generate $\bm{x}_i$. Then, for any rotation matrix $\bm{P}$ we have:
\begin{align*}
\bx_i 
= \bm{B}\bz_i + \bm{\varepsilon}_i 
= \bm{B}\bm{P}\bm{P}^\top\bz_i + \bm{\varepsilon}_i
= \widetilde{\bm{B}}\widetilde{\bz}_i + \bm{\varepsilon}_i, \quad \text{where }\widetilde{\bm{B}}=\bm{B}\bm{P} \text{ and } \widetilde{\bm{z}}_i=\bm{P}^\top \bm{z}_i.
\end{align*}
The above reparameterization is valid as the standard Gaussian is invariant under rotation: $\widetilde{\bm{z}}_i\sim \mathcal{N}_K(\bm{0}, \bm{I})$. Consequently, the model is not identifiable. This lack of identifiability is problematic for interpreting groups of features via the loadings matrix $\bm{B}$: There are an infinite number of loadings matrices with the same likelihood; based on the data alone, we cannot determine the true loadings matrix.

\begin{figure}
\centering
    \includegraphics[width=0.8\textwidth]{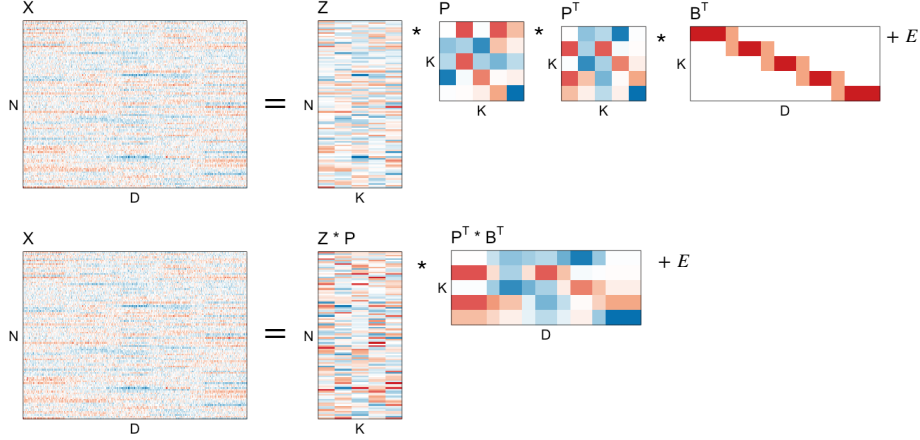}
    \caption{Linear factor analysis is not identifiable; applying any rotation matrix $\bm{P}$ leads to another model with equivalent likelihood.}
\end{figure}

This identifiability problem can be resolved by placing further assumptions on the data generating process. Typically, further assumptions are placed on either (i) the class of loadings matrices, (ii) the distribution $\zpr$,  or both.  The goal is to specify assumptions on $\bm{z}_i$ and $\bm{B}$ that are not preserved under rotation. Then, $(\bm{P}^\top\bm{z}_i, \bm{B}\bm{P})$ will not satisfy these assumptions for any nontrivial rotation $\bm{P}$, and thus will not be equivalent solutions.

\begin{remark}
In addition to the identifiability issues from rotational invariance, there are identifiability issues regarding the noise variance, $\bm{\Sigma}$. Specifically, further restrictions are required to uniquely determine $\bm{\Sigma}$ from the marginal covariance of the data, $\bm{B}\bm{B}^\top + \bm{\Sigma}$ \citep{anderson1956statistical,fruhwirth2023counts}. Examples of such restrictions include the row-deletion property \citep{anderson1956statistical}, the counting rule \citep{fruhwirth2023counts}, or imposing additional constraints on $\bm{B}$ \citep{bing2020adaptive}. These restrictions are imposed to uniquely determine the elements of $\bm{B}$ and $\bm{\Sigma}$ (up to sign/column perturbations); however,  in some cases, we may be interested in weaker notions of identifiability, such as identifying only the support of $\bm{B}$.
\end{remark}

\subsection{Identifiability assumptions on the loadings matrix}

\noindent
\parhead{Triangular loadings matrices.} One early approach to identify the linear factor analysis model was to constrain the class of loadings matrices to be lower triangular matrices with positive diagonal elements (PLT) \citep{anderson1956statistical}. There are no rotation matrices that preserve this PLT structure except the identity, which limits the set of equivalent-likelihood solutions.  The PLT condition is commonly used in the statistical factor analysis literature \citep{carvalho2008high,fruhwirth2010parsimonious}.

An issue with the PLT condition, however, is that it ``fixes'' the factors to the first $\zdim$ dimensions of $\bm{X}$. That is, with the triangular assumption, the generative model is:
\begin{align*}
    x_{i1} &= b_{11}z_{i1} + \varepsilon_{i1}, \\
    x_{i2} &= b_{21}z_{i1} + b_{22}z_{i2} + \varepsilon_{i2} \\
    &\vdots \\
    x_{i\zdim} &= b_{\zdim 1}z_{i1} + \cdots b_{\zdim \zdim}z_{i\zdim} + \varepsilon_{iK}, \quad i = 1,\dots, N.
\end{align*}
The first factor dimension $\bm{z}_{\cdot 1}$ then corresponds to the first observed feature $\bm{x}_{\cdot 1}$; the second factor dimension corresponds to a linear combination of the first two observed features, and so on. This correspondence may be problematic in applications where the first $\zdim$ observed features do not have special status for defining the factors.

More recently, \citet{fruhwirth2023counts} generalized the PLT condition to the unordered generalized lower triangular (UGLT) condition, proving that the UGLT condition resolves unidentifiability due to rotational invariance. The UGLT condition allows for an unknown set of $\zdim$ (out of $\xdim$) features to have special status in defining the factors (instead of the first $\zdim$ features, as in the PLT).

\noindent
\parhead{Anchor features.} \citet{bing2020adaptive} propose learning which observed features act as ``anchors'' for the factor dimensions. Their method identifies the loadings matrix under the key assumption that each factor has at least two ``pure variables'', or what we call ``anchor features''. An anchor feature for a particular factor is an observed feature that depends only on that factor. Anchor features do not need to be known in advance; they can be inferred from the data covariance, under additional assumptions.  See \Cref{fig:correlation} for a visualization—anchor features exhibit higher mutual covariance than non-anchor features.

The anchor feature assumption means that the true loadings matrix $\bm{B}$ contains two submatrices that are the identity (after row and column permutation):
\begin{align}
\bm{B}^\top=
\begin{pmatrix}
\bm{I}_{\zdim\times \zdim} &
\bm{I}_{\zdim\times \zdim} &
\widetilde{\bm{B}}^\top
\end{pmatrix}. \label{eq:pure-variables}
\end{align}
This structure on $\bm{B}$ again eliminates all non-trivial rotations because this particular sparsity pattern is not invariant to rotation. 

The anchor feature assumption has also been utilized for identification in related latent variable models. For example, this is analogous to the ``anchor word'' assumption in topic models \citep{arora2013practical}. This is also related to the separability condition \citep{donoho2003nmf} in the non-negative matrix factorization (NMF) literature.

The anchor feature and separability assumptions also help explain the empirical success of sparse Bayesian factor analysis. In Bayesian factor analysis, sparsity-inducing priors are often placed on the loadings matrix $\bm{B}$; these sparse models often yield interpretable groups of features in real data settings \citep{knowles2011nonparametric,bhattacharya2011sparse,rovckova2016fast,wang2021empirical}. Such priors place higher probability on sparse matrices, which are more likely to satisfy the anchor assumption than dense matrices.

\subsection{Identifiability assumptions on the distribution of the factors}

\noindent
\parhead{Non-Gaussian independent factors.} In psychometrics and statistics, a popular approach to factor analysis is to first calculate principal components or maximum likelihood estimates, and then apply a rotation such as Varimax \citep{kaiser1958varimax}. (See for example,  multivariate analysis textbooks such as \citet{anderson2003,jolliffe2002principal}.) 
However, this approach lacked theoretical justification until recently, when \citet{rohe2023vintage} proved that PCA and Varimax consistently estimates the latent factors (provided the factors have a particular heavy-tailed distribution).  Intuitively,  Varimax rotates factors toward non-Gaussian solutions. As non-Gaussian distributions of independent variables are not rotationally invariant (as known from Maxwell's theorem,  \citealp{maxwell1860illustrations}; see also III, 4 in \citealp{feller1971introduction}), this assumption removes indeterminances due to rotational invariance.

The relationship between non-Gaussianity and identifiability in factor models is also widely exploited in the literature on independent component analysis \citep[ICA,][]{comon1994independent,eriksson2004identifiability}. Central to this literature is the Darmois-Skitovich lemma and its extensions, which extends Maxwell's characterization of the normal distribution in terms of independent linear forms. See \citet{kagan1973} for a detailed account and related results. An application to factor analysis is the following widely known identifiability result for ICA: If $\bx_i = \bm{B}\bz_i$, where $\bz_i$ consists of independent non-Gaussian components, then $\bm{B}$ is identifiable up to permutation and scaling. 
This result crucially relies on non-Gaussianity, as even two Gaussian components is enough to break identifiability.

\noindent
\parhead{Correlated factors.} Typically  the latent factors are assumed to be uncorrelated, but recent work has begun to investigate the case where factors are correlated: \citet{bing2020adaptive} allow for correlated factors, as long as the variance of each factor is greater than its covariance with any other factor; \citet{rohe2023vintage} implicitly allow for correlated factors by estimating a ``middle matrix'' between the loading and the factor matrices.

\subsection{Beyond classical factor analysis}
In the machine learning literature, similar identifiability conditions have been proposed using the language of graphical models and ICA. For example: 
\begin{itemize}
    \item The PLT assumption is equivalent to assuming a known causal ordering on the factors \citep[e.g.][]{germain2015made,shen2022weakly};
    \item Anchor features are equivalent to assuming a particular sparsity pattern on the graphical model implied by the loadings matrix $\bm{B}$ \citep[e.g.][]{silva2006learning,arora2013practical}; \item Correlated factors arise naturally with causally related latent factors, and assumptions on correlated factors are closely related to causal restrictions on the latent factors \citep[e.g.][]{silva2006learning}.
\end{itemize}
\noindent
\citet{silva2006learning} 
made the connection between causality and factor models explicit by presenting a framework for learning causal structure in linear latent variable models. For example, can we determine which biological processes are \emph{causally} related? In essence, this is the goal of CRL: How do we learn latent factors (i.e. representations), \emph{and} learn the causal relationships between them? 
To study such questions, we need concepts from the graphical approach to causality \citep{pearl2009causality}, which we now review.

\section{Causal Models and Latent Variables}
\label{sec:causal}

Evidently, there is a close connection between classical assumptions in factor analysis and graphical assumptions found in machine learning and causal inference. In fact, the factor analysis model is equivalent to a graphical model known as a measurement model,
where the observed variables $\bx_{i}$ are conditionally independent given latent variables $\bz_{i}$. 
Measurement models originate from the psychometric literature on confirmatory factor analysis, where the causal relationships among latent variables are typically specified beforehand \citep[e.g.][]{bentler1980multivariate,bollen1989structural}.
By contrast, the contemporary measurement model used in CRL---where both the latent variables and their causal structure are unknown and must be inferred from data---was first explicitly formulated by \citet{silva2006learning}. 
Although this early work focused on linear models, modern advances increasingly emphasize nonlinear models.

{For context, we provide some formal background on these notions in this section.
In doing so, we will also present some} minimal background on graphical models, emphasizing linear models for straightforward comparison with the factor analytic frameworks discussed above. However, readers familiar with graphical models will recognize that these concepts immediately generalize to nonlinear models.

\subsection{Graphical approach to causality}
\label{sec:causal:graphical}

The graphical approach to causality is built upon the fundamental concept of a Bayesian network. A Bayesian network (BN) over a random vector $\bx_{i}=(x_{i1},\ldots,x_{i\xdim})\in\R^{\xdim}$ with joint distribution $P$ is any directed acyclic graph (DAG) $\bn=(\ver,\edg)$ with $\ver=\bx_{i}$ such that $P$ factorizes over $\bn$, i.e. it satisfies the \emph{Markov condition} or \emph{is Markov to $\bn$}:
\begin{align}
\label{defn:markov}
P(\bx_{i})
=\prod_{j=1}^{\xdim}P(\bx_{ij}\given \pa_{\bn}(j)).
\end{align}
The dependence on $\bn$ in the above expression arises from the parent sets $\pa_{\bn}(j):=\{k : \bx_{ik}\to \bx_{ij}\text{ in }\bn\}$. As is conventional in graphical models, we will often simply write $k$ in place of $\bx_{ik}$, e.g. $k\to j$ indicates an edge pointing from $\bx_{ik}$ to $\bx_{ij}$.
The purpose of the Markov condition is to connect distributions and graphs: 
Although every distribution is Markov to \emph{some} $\DAG$, the Markov condition allows us to define statistical models through structural assumptions on distributions.

We are interested in models that contain 
both observed and hidden components: $(\bx_{i},\bz_{i})=(x_{i1},\dots, x_{i\xdim}, z_{i1},\dots, z_{i\zdim})$. In this case, the $\DAG$ $\bn$ and the factorization \eqref{defn:markov} extends over the joint random vector $(\bx_{i},\bz_{i})$. Without some type of structural assumption on $\bn$, the types of structures that can be represented by a Bayesian network with latent variables are known to be quite exotic, and have been the subject of much research \citep{richardson2017nested,evans2018margins}.
We will focus on the case of factor structures, defined below, in order to emphasize the connection with factor analysis, although one could of course study more exotic latent structures.

Factor models as in (\ref{eq:factor-analysis}) implicitly impose structural assumptions in terms of the $\DAG$ $\bn$ \citep{drton2007algebraic}. That is, if $\bn$ is a BN for a factor model over $(\bx_{i},\bz_{i})$, then $\bn$ has a factor structure defined as follows:
\vspace{0.5em}
\begin{definition}
\label{defn:factor}
    A DAG $\bn$ is said to have a \emph{factor structure} if there are no edges between observed variables or from observed variables to latent variables. A distribution over $(\bx_{i},\bz_{i})$ is said to follow a \emph{factor model} if $(\bx_{i},\bz_{i})$ is Markov to a DAG $\bn$ with a factor structure.
\end{definition}
\noindent
Factor models are also known as \emph{measurement models};
for consistency of presentation, we stick to the term \emph{factor structure}, acknowledging that this is synonymous with earlier usage of \emph{measurement model}.
This definition implies that $\bx_{i}$ are conditionally independent given $\bz_{i}$, as in a factor model. 
In fact, if we assume linear relationships and Gaussian noise, Definition~\ref{defn:factor} is equivalent to (\ref{eq:factor-analysis}). 
Figure~\ref{fig:intro} depicts two prototypical examples of a factor model.

\subsection{Latent causal graphs}
\label{sec:causal:latent}
\begin{figure}[t]
\centering
\begin{tikzpicture}[
    vertex/.style={circle, draw, minimum size=20pt},
    zlabel/.style={vertex, fill=white},
    xlabel/.style={vertex, fill=gray!20}
]

\foreach \i in {1,2,3} {
    \node[zlabel] (z\i) at (2*\i-4,2) {$z_{\i}$};
}

\path (z1) -- (z2) coordinate[midway, yshift=30pt] (m12);
\path (z1) -- (z3) coordinate[midway, yshift=30pt] (m13);
\path (z2) -- (z3) coordinate[midway, yshift=30pt] (m23);

\foreach \i in {1,...,7} {
    \node[xlabel] (x\i) at (1.5*\i-6,0) {$x_{\i}$};
}

\node[draw, rounded corners, very thick, fit=(z1) (z2) (z3) (m12) (m13) (m23), inner sep=10pt] (box) {};

\node[anchor=north west] at ($(box.north west)+(5pt,-5pt)$) {$\latgr$};

\draw[->, dashed] (z1) -- (x1);
\draw[->, dashed] (z1) -- (x2);
\draw[->, dashed] (z1) -- (x3);
\draw[->, dashed] (z2) -- (x3);
\draw[->, dashed] (z2) -- (x4);
\draw[->, dashed] (z2) -- (x5);
\draw[->, dashed] (z3) -- (x5);
\draw[->, dashed] (z3) -- (x6);
\draw[->, dashed] (z3) -- (x7);

\draw[->, very thick, bend left=45] (z1) to (z2);
\draw[->, very thick, bend left=45] (z1) to (z3);
\draw[->, very thick, bend left=45] (z2) to (z3);

\end{tikzpicture}
\caption{Decomposition of the full causal factor model $\bn$ over $(\bx_i,\bz_i)$ into a latent causal graph $\latgr$ over $\bz_i$ and a bipartite graph (dashed arrows) from $\bz_i$ to $\bx_i$.}\label{fig:decomp}
\end{figure}
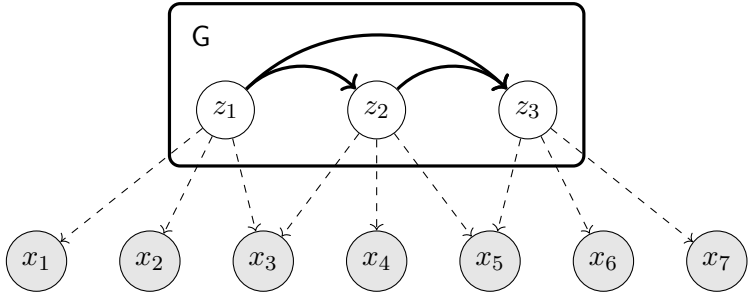

The factor model makes explicit
that we are not interested in causal relationships \emph{between} observables---it is not $\bx_{i}$ itself that we are interested in modeling causally, but rather the latent structure over $\bz_{i}$. 
The primary causal relationships of interest involve only the hidden variables. 
Thus, all of the modeling emphasis is shifted over to the latent factors. 
The goal is to use $\bx_{i}$ to \emph{infer} a latent causal structure: This involves learning both the latent variables $\bz_{i}$ and their causal relationships. 

More formally, factor models always decompose into two parts (\Cref{fig:decomp}): A bipartite component that involves only directed edges from $\bz_{i}\to \bx_{i}$, and a \emph{latent causal graph} $\latgr$ that involves only directed edges between the latent variables $\bz_{i}$. Thus, the \emph{full} causal graph over $(\bx_{i},\bz_{i})$ jointly is denoted by $\bn$, whereas the \emph{latent} causal graph over $\bz_{i}$ only is denoted by $\latgr$. We omit reference to the bipartite graph since this is not the main focus.

\begin{figure}[t]
    \centering
    \begin{minipage}[c]{0.22\textwidth}
\begin{tikzpicture}[>=Stealth, node distance=1.25cm, transform shape, scale=0.9]
\node (Z1) [draw, circle] {$z_1$};
            \node (Z2) [draw, circle, right=of Z1] {$z_2$};
            \node (Z3) [draw, circle, below=of Z2] {$z_3$};
            \node (Z4) [draw, circle, left=of Z3] {$z_4$};
\draw[->] (Z1) -- node[above] {$2$} (Z2);
            \draw[->] (Z1) -- node[above right] {$-3$} (Z3);
            \draw[->] (Z2) -- node[right] {$3$} (Z3);
            \draw[->] (Z3) -- node[below] {$1$} (Z4);
        \end{tikzpicture}
    \end{minipage}\begin{minipage}[c]{0.65\textwidth}
        \begin{align*}
        \left.
        \begin{aligned}
        z_{i1} &= \nu_{i1} \\
        z_{i2} &= 2z_{i1} + \nu_{i2} \\
        z_{i3} &= -3z_{i1} + 3z_{i2} + \nu_{i3} \\
        z_{i4} &= z_{i3} + \nu_{i4}
        \end{aligned}
        \right\}
        \Rightarrow
        \bz_{i}
        = \latsem\bz_{i}+\bm{\nu}_i,
        \quad
        \latsem
        = \begin{pmatrix}
        0 & 0 & 0 & 0 \\
        2 & 0 & 0 & 0 \\
        -3 & 3 & 0 & 0  \\
        0 & 0 & 1 & 0 \\
        \end{pmatrix}
        \end{align*}
    \end{minipage}\caption{An example of a linear structural equation model over latent variables $\bz_{i}$.}
    \label{fig:linsem}
\end{figure}
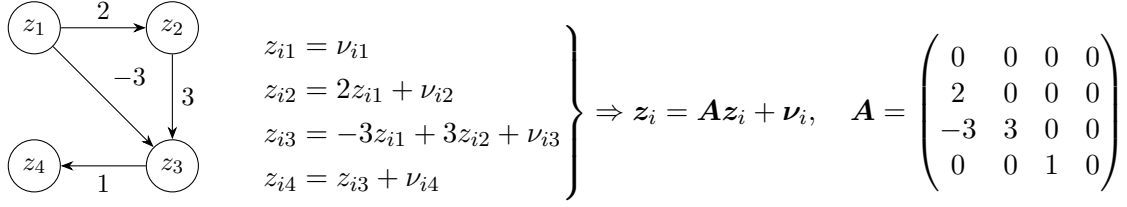

In the linear case, also known as a linear structural equation model \citep[SEM,][]{bollen1989structural,pearl2009causality}, the model for $\zpr(\bz_{i})$ becomes
\begin{align}
\label{eq:lingam}
\bz_{i}
= \latsem\bz_{i} + \bm{\nu}_i,
\quad
\latsem=(\latsemwt_{jk})\in\R^{\zdim\times \zdim},
\end{align}
where $\bm{\nu}_i$ are jointly independent, zero-mean random variables. Then we can associate to $\bz_{i}$ a latent causal graph $\latgr$ by reading off the nonzero entries of the matrix $\latsem$: Simply define $\latgr=(\ver,\edg)$ by
\begin{align}
\label{eq:lingr}
\ver
= \bz_{i},
\quad
\edg
= \{(j,k) : \latsemwt_{jk}\ne 0\}. 
\end{align}
See \Cref{fig:linsem} for an example.
It turns out that as long as $\latgr$ is acyclic, then $\latgr$ is a BN for $\bz_{i}$.  In other words, the matrix $\latsem$ explicitly encodes a BN for $\bz_{i}$.

\subsection{Interventions and causality}
\label{sec:causal:ivn}

The edges in a DAG can be interpreted as indicating the causal dependencies in a system: The edge $z_{ik}\to z_{ij}$ indicates that $z_{ij}$ depends causally on $z_{ik}$ in the sense that changes to $z_{ik}$ propagate to $z_{ij}$. Returning to the linear SEM example in \eqref{eq:lingam},  such causal dependencies can be seen by writing the model component-wise:
\begin{align}
z_{ij}
= \sum_{k=1}^{K} \latsemwt_{jk}z_{ik} + \nu_{ij}.
\label{eq:linear-sem}
\end{align}
Explicitly, $z_{ij}$ is a noisy function of its parents.

\noindent
\parhead{Interventions and independent mechanisms.} 
This causal interpretation allows us to reason about how the system changes when we externally manipulate the values of certain variables, i.e. an intervention. An intervention occurs when we fix the value of a certain variable (or set of variables) to a particular value. When this happens, the causal parents of the intervened variable are ignored, akin to deleting incoming edges to this node.

To make this concrete, consider the linear SEM in \Cref{fig:linsem}.  Without any external interventions, the value of $z_{i3}$ is determined by the values of $z_{i1}$ and $z_{i2}$. The value of $z_{i2}$, in turn, is determined by $z_{i1}$, whereas $z_{i1}$ is an independent exogenous variable that is not determined by any other $z_{ik}$. Now suppose we intervene on $z_{i2}$: Then the edge from $z_{i1}\to z_{i2}$ is broken, that is, the value of $z_{i1}$ is no longer relevant for determining $z_{i2}$: The value of $z_{i2}$ is fixed by external manipulation. This then propagates down to $z_{i3}$, whose value is determined by the (unintervened) value of $z_{i1}$ and the intervened value of $\bz_{i2}$. The result is a \emph{new} linear SEM
\begin{align*}
z_{i1} &= \nu_{i1} \\
z_{i2} &= z_{i2}^{*} \\
z_{i3} &= -3z_{i1} + 3z_{i2}^{*} + \nu_{i3} \\
z_{i4} &= z_{i3} + \nu_{i4},
\end{align*}
where $z_{i2}^{*}$ is the value that $z_{i2}$ is manipulated to. To denote the new distribution of $z_{i3}$ given this intervention on $z_{i2}$, the convention is to use do-notation:
\begin{align*}
P(z_{i3}|z_{i1}, \mathrm{do}(z_{i2}=z_2^*)).
\end{align*}

Implicit in this formalization of interventions is the principle of \emph{modularity} or \emph{independent mechanisms}: Intervening on a node changes the local conditional distribution of that node, but leaves the remaining conditional distributions unchanged. See \citet{peters2017elements} (Chapter 2) for more discussion on the principle of independent mechanisms. 

\noindent
\parhead{Latent interventions.} 
The factor model (\Cref{defn:factor}) makes explicit that intervening on the observed features $\bx_{i}$ is not meaningful.
Instead, we are interested in intervening on \emph{latent variables}, which might seem strange at first glance. After all, we do not directly observe latent variables. In many applications, however, this is quite natural: recall the motivating examples in \Cref{sec:motiv}.

When the observed features are pixels (\Cref{fig:intro-example-2}), intervening on individual pixels is not meaningful outside of toy applications. 
In practice, however, we can obtain additional images or data augmentations in which the object of the image is translated or rotated, or the background is changed. Each of these changes can be viewed as an intervention. However, it is difficult to model such interventions at the pixel level; e.g. in different images,  different subsets of pixel values correspond to the background.~A solution to this problem is to learn a latent representation $\bm{z}_i$ of each image $\bx_{i}$ where the dimensions $z_{ik}$ have a consistent interpretation for all $i$; e.g. $z_{ik}$ corresponds to the rotation of the images. This type of data is readily available in applications, and can even be easily generated with the aid of simulators \citep[e.g.][]{von2021self,brehmer2022weakly}.

In genomics, the observations are gene expression levels (\Cref{fig:intro-example-1}). Genes may coordinate as part of a biological pathway; collectively, these genes may induce effects in other variables. As such, it may be more efficient to aggregate these gene subsets into single causal variables representing gene programs or biological processes. Other examples of latent interventions include medical treatments (drugs, surgeries), radiation exposure, or other epigenetic factors (DNA methylation, histone modification, environmental factors).

We refer to an intervention where a single dimension of $\bm{z}_i$ is intervened upon as a \emph{single-node intervention}. For a linear SEM, we formalize this as follows.
\vspace{0.5em}
\begin{definition}[Single-node latent interventions]\label{def:single-node}
Given an intervention $e\in\mathcal{E}$, denote the resulting distribution on $\bm{z}$ as $\zinv$.
An interventional distribution $\zinv$ is a single-node intervention of the latent distribution $\zpr$ if there is a single target node $t(e) \in [\zdim]$, and the assigning equation of $z^{(e)}_{i,t(e)}$ is:
\begin{align}
z_{i,t(e)}^{(e)} &= \sum_{k=1}^K a^{(e)}_{t(e),k} z_{ik}^{(e)} + \nu_{i,t(e)} + c^{(e)} 
\end{align}
while leaving all other equations $z_{ij}$ for $j\neq t(e)$ unchanged. We assume $\latsemwt_{t(e),k}^{(e)} \neq0$ only if $k\to t(e)$ in $\gr$; that is, no parents are added. The variable $c^{(e)}$ denotes a shift of the mean; the variance of the noise $\nu_{i,t(e)}$ may also be different to that of $\zpr$. 
An intervention is called perfect if $\latsemwt_{t(e), k}^{(e)} =0$ for all $k\in [K]$, i.e. all connections to former parents are removed. 
\end{definition}

Intervention targets can be either \emph{known} or \emph{unknown}: When intervention targets are known, there is no ambiguity about the mapping $t(e)$ defined above. A considerably more challenging setting arises when intervention targets $t(e)$ are unknown. In CRL, since the intervention targets are latent and unobserved, these interventions are also unknown---in fact, we may not even know the correct number of latent factors to begin with. 

This leads to additional complications in CRL: Although there is a growing literature on learning under unknown interventions \citep{eaton2007exact,kocaoglu2019characterization,jaber2020causal,squires2020permutation,castelletti2022network,hagele2022bacadi,faria2022differentiable}, this line of work focuses on interventions on measured variables (i.e. $x_{ij}$). 
When interventions are unknown \emph{and} latent (i.e. $z_{ij}$), much less is known—this is a developing area of research driven by problems in CRL; e.g. \citet{jiang2023learning}.

\section{What is causal representation learning?}
\label{sec:crl}

With these preliminaries on factor models and causality out of the way, we are ready to discuss CRL in more precise terms.  Simply put, CRL takes a modern view on factor analysis, driven by the success of generative AI and guided by three core principles as outlined in the introduction: Flexibility, interpretability, and transferability.
\begin{itemize}
    \item \emph{Flexibility} is achieved by allowing (a) general nonlinear relationships through arbitrarily complex neural networks, as well as (b) correlations between latent variables;
    \item \emph{Interpretability} is achieved by enforcing sparsity (to simplify the model) and causality (to allow for an interventionist interpretation of the latent factors) through the use of causal graphical models;
    \item \emph{Transferability} is achieved by using causal models, which are widely believed to yield stable, invariant, transferable predictions across changing contexts and environments.
\end{itemize}

To capture these three core principles, CRL begins with a general nonlinear factor model of the form $\bm{x}_i = f(\bm{z}_i) + \bm{\varepsilon}_i$, 
where the latent $\bm{z}_i$ are related by some latent causal graph $\latgr$ and the nonlinearity $f$ is typically estimated with deep neural networks, as in a deep generative model, and causal relationships are inferred through the use of data arising from multiple environments (e.g. across different experimental interventions, time points, etc.).
Examples of generative models explicitly of this form include diffusion models, variational autoencoders, and flows.
This class of models continues to dominate leaderboards tracking progress in generative AI~\citep{hugging2025text,liu2024evalcrafter}.

\subsection{Problem definition}

To discuss issues such as identifiability, we now formalize the CRL problem. Given a function class $\fcns$ and a family of prior distributions $\zprs$, we assume the observed data $\bm{x}_i\in \mathcal{X}$ is generated from the latent $\bm{z}_i\in \mathcal{Z}$ via an unknown mixing function $f \in \fcns$:
\begin{align}
\bm{x}_i = f(\bm{z}_i) + \bm{\varepsilon}_i, \quad \bm{\varepsilon}_i\stackrel{\textup{iid}}{\sim} p(\bm{\varepsilon}),\quad i = 1,\dots,n,  \label{eq:nonlinear-rep-learning}
\end{align}
where the latent $\bm{z}_i\sim\zpr \in \zprs$ are generated according to a linear structural equation model:
\begin{align}
\bm{z}_i = \latsem\bm{z}_i + \bm{\nu}_i, 
\label{eq:linear-sem-crl}
\end{align}
where $\latsem$ is the SEM matrix for some DAG $\gr$ (cf. \eqref{eq:lingam}) and $\bm{\nu}_i$ are jointly independent, zero-mean random variables. These assumptions are not necessary, but will help simplify the discussion here; see Remark~\ref{rem:sem} below.

The goal is to infer the unknown function $f$ and the prior on the latents $\zpr$ including $\latsem$, which encodes the causal graph $\latgr$ of $\bz$. The  parameters $(f, \zpr, \latsem)$ are important for sampling from the model and understanding the effect of interventions on the observed data. In CRL, the motivation for learning these parameters is interpretability (via sparse structure in $\latsem$ or $f$) and transferability to new domains (via the causal mechanisms encoded by $\latsem$).   In practice, it may be enough to identify certain functionals of the parameters $(f, \zpr, \latsem)$, such as a subgraph or the behavior of $f$ on subspaces of interest.

These parameters are specified by the class of transformations $\fcns$ and the family of distributions $\zprs$ over $\bz$. For example, in the linear factor analysis model $\fcns$ is a class of linear maps and $\zprs$ is typically the family of Gaussian distributions. The overall model can be formalized as the collection of distributions induced by these families. More general models can be obtained by expanding $\fcns$ to include nonlinear transformations or $\zprs$ to include non-Gaussian distributions. 
Since the only difference between \eqref{eq:nonlinear-rep-learning} and the earlier linear factor model \eqref{eq:factor-analysis} is the replacement of the linear map $\bm{B}\bm{z}_i$ with the nonlinear map $f(\bm{z}_i)$, the former is a special case of the latter. Moreover, since even linear factor models are not identifiable, it follows that \eqref{eq:nonlinear-rep-learning} is also not identifiable. A key goal of CRL is therefore to determine conditions under which this model becomes identifiable and estimable.

\begin{remark}
\label{rem:dnn}
    The model \eqref{eq:nonlinear-rep-learning} is an assumption on the distribution that can be made independently of whether neural networks are used to fit the data or not; in particular it is not necessary to assume that $f$ \emph{itself} is a neural network. A common choice for $\fcns$ is the family of invertible, differentiable functions (i.e. diffeomorphisms); see \citet{kivva2022identifiability} for a detailed discussion of the implications of various functional restrictions. Of course, in practice we are interested in using neural networks to fit this model nonparametrically.
\end{remark}

\begin{remark}
\label{rem:sem}
We focus on models where the latents satisfy a linear structural equation model and the noise is additive, although these assumptions are not necessary. The literature has studied more general models \citep[e.g.][]{jiang2023learning,varici2023score,varici2024general,von2023nonparametric}; see the end of Section~\ref{sec:nonlinear} for a brief overview. 
\end{remark} 

\begin{remark}
\label{rem:nonlinearfa}
The model \eqref{eq:nonlinear-rep-learning} is a nonlinear factor model, which has a long history in both statistics and ML; see \cite{amemiya2001fa} for a review. See also \cite{zhou2024bayesian,xu2023identifiable} for recent developments. Although the mathematical models share similarities, what differentiates CRL from prior work on nonlinear factor analysis is the focus on deep learning and causality that drives much of CRL.
For example, a key distinction arises from the latent SEM \eqref{eq:linear-sem-crl} and the causal, interventionist semantics it provides.
\end{remark}

\subsection{Generation vs. understanding}

As noted in the introduction, in applications of generative AI there is a crucial distinction between the problems of \emph{generation} and \emph{understanding} that parallels the two cultures of statistical modeling. 
Since the vast majority of the literature on generative AI focuses on generation only, we pause for a moment to outline in more detail these two goals:
\begin{enumerate}
\item \emph{Generation} (density estimation). Here, we evaluate a model based on its ability to reconstruct the true data density of $\bm{x}_{i}$, and ignore what latents are learned or how they can be interpreted. More formally, generation quality is measured by distributional metrics such as total variation, Wasserstein distance, or Kullback-Liebler (KL) divergence.
For example, if $p$ represents the density of $\bm{x}_{i}$, then we can evaluate the quality of a model $\wh{p}$ with the KL divergence $\textup{KL}(p\,\Vert\,\wh{p})$. Closely related to this goal is sampling from $p$, which is a key objective in  many applications. 
\item \emph{Understanding} (parameter estimation). Here, we evaluate a model based on the quality of the latents themselves, which are captured by the model parameters $\bm{z}_{i}$ and $f$. This could mean recovering the precise latent variables $\bm{z}_{i}$ for each sample $\bm{x}_{i}$, the latent causal graph $\gr$, or even just the distribution over $\bm{z}_{i}$. In other words, our model should capture meaningful semantics about the generative process.
\end{enumerate}
The distinction between these two goals has two important statistical implications: 1) Generation has progressed remarkably without understanding the underlying models, suggesting generation is more tractable than understanding; and 2) To understand models, parameters should be identifiable so that they can be consistently estimated as in traditional statistical models.

\subsection{Identifiability} \label{sec:crl-identifiability}

It is clear that \eqref{eq:nonlinear-rep-learning} is not identifiable: Different parameters $(f,\zpr)$ lead to the same distribution $p(\bx_{i})$ over the observables. For generation, identifiability is not necessary, however, as we have discussed, we want to go beyond generation to deeper questions of understanding and interpretability. Statistically, we would like to discuss parameter estimation in \eqref{eq:nonlinear-rep-learning}, analogous to how factors are estimated in classical linear factor models. 
While standard in the statistics literature, the relevance of identifiability for ML/AI applications has also been articulated in recent work \citep{damour2020underspecification,reizinger2024position}.

We shall distinguish between two types of identifiability in the model \eqref{eq:nonlinear-rep-learning}:
\begin{enumerate}
    \item \textbf{Statistical identifiability} of the factor model parameters $(f,\zpr)$.
    \item \textbf{Causal identifiability} of the causal graph $\latgr$ and its parameters (e.g. $\latsem$).
\end{enumerate}
Statistical identifiability can be understood in analogy with factor analysis; i.e. identifiability of the latent factors. Causal identifiability is the distinct problem of identifying the causal relationships between the latent factors as encoded by the latent causal graph $\latgr$.  

These notions of identifiability are distinct: Statistical identifiability does not imply causal identifiability, and vice versa. That $\latgr$ cannot be identified from $\zpr$ is a well-known result in graphical models: 
Simply by the chain rule of probability, $\zpr$ can be factorized over different graphs, each of which encodes (potentially) different conditional independence statements. This leads to the notion of the \emph{Markov equivalence class}, which collects all the factorizations of $\zpr$ that encode exactly the same conditional independence relations as $\zpr$.
For example, consider two dependent variables $(z_{i1},z_{i2})$ with joint distribution $\zpr(z_{i1},z_{i2})$. By the chain rule of probability, $\zpr$ admits two factorizations:
\begin{align*}
\zpr(z_{i1})\zpr(z_{i2}\,|\,z_{i1}) \quad \text{and} \quad \zpr(z_{i2})\zpr(z_{i1}\,|\,z_{i2}).
\end{align*}
That is, $\zpr$ factorizes with respect to both the graph $z_{i1}\to z_{i2}$ and $z_{i2} \to z_{i1}$. Consequently, $\gr$ is not identifiable from $\zpr$ (for more on Markov equivalence, see \citealp{lauritzen1996graphical,pearl2009causality}). The converse—that $\zpr$ cannot be identified from $\latgr$—is also classical; see \citet{drton2023algebraic} for recent developments.

In practice, both notions of identifiability are desirable, however, we may be limited to one or the other depending on the available data and assumptions. Whether or not we are interested in causal or statistical identifiability depends on the problem at hand, although in typical applications causal identifiability comes after statistical identifiability; i.e. once we can identify and interpret the representations, we are interested in the effects of manipulating these representations.

\subsection{Identifiability vs. interpretability}
\noindent 
While identifiability is an important prerequisite for statistical estimation and reliability, it is not sufficient for interpretability. This is well-known: For example, the latent factors recovered by PCA are well-defined but often lack clear interpretation. Here it is helpful to distinguish between two layers of interpretability: First, \emph{can these factors be interpreted at all?} and second, \emph{if they can be interpreted, do they correspond to meaningful concepts?} While identifiability partially addresses the first layer, it does not address the second layer.

Moving beyond identifiability, the first layer can also be addressed by leveraging structure in $f$: When $f$ is identifiable, latent factors may be understood by how they relate to the observed features. In linear models, this is reflected in the loadings matrix: If a group of observed features with shared meaning (e.g. genes) load on the same factor, that factor can inherit this interpretation (e.g. the biological function of these genes). Recent work extends this idea to nonlinear settings \citep{ainsworth2018oivae,moran2022identifiable}.

The challenge with the second layer is that there is no formal definition of ``meaningful'': What is meaningful depends on context and is subjective. For example, size, colour, and shape are nearly universal concepts, however, their \emph{precise} definitions can vary from one application to the next; e.g. ``size'' can be interpreted as physical size, file size, or population size. Each of these interpretations measures some aspect of ``largeness'', but in incomparable units for different applications.

Thus, to capture the second layer, we often use auxiliary information in the form of concept labels, metadata, or environment annotations that capture this subjective information. This allows us to ask that not only we identify the latent factors and the causal graph, \emph{but also that the factors meaningfully capture subjective qualities}. This adds an additional constraint on the CRL problem that connects with previous work on interpretable ML \citep{koh2020concept,schut2023bridging}, with the main difference being the added focus on identifiability in CRL. 
For example, to identify meaningful concepts, \citet{taeb2022provable} require the factors to predict an auxiliary label; \citet{zhang2023identifiability} learn a mapping from intervention label to the latent space; and \citet{rajendran2024learning} use concept-conditional environments and contrastive learning to learn interpretable concept representations.
These ideas also relate to an emerging
literature that explores how to identify human-interpretable concepts in foundation models (such as large language models) that is discussed in more detail in Section~\ref{sec:conc} (under ``Large Language Models''). As this literature is nascent and rapidly developing, we defer a detailed discussion to the papers themselves.

\noindent
\parhead{Confounding.}
So far we have assumed every latent factor is of interest. In the literature, latent factors have often been interpreted as confounding variables of no inherent interest \citep[e.g. batch effects,][]{wang2019blessings,cevid2020spectral}. It is to be expected that there will be both spurious and meaningful latent factors in data; how to handle and understand these factors is an open question in CRL. Some promising directions in this regard include: allowing both dense and sparse factor loadings \citep{chandrasekaran2011rank,zhao2016bayesian,moran2021spike}, and utilizing interventional data \citep{luo2025gene}.

\section{Identifiability in causal representation learning}
\label{sec:nonlinear}

In this section, we consider in detail the identifiability of the model (\ref{eq:nonlinear-rep-learning}-\ref{eq:linear-sem-crl}). First, we discuss approaches to statistical identification of the model parameters $(f, \zpr)$. Then, we discuss approaches to causal identification, i.e. learning the causal graph between the latent variables using interventions.

\subsection{Statistical identifiability: Nonlinear representation learning}

In \Cref{sec:linear}, we reviewed the identification problem in linear factor analysis. Of course, identification becomes harder if we relax linearity: Informally, we can always re-write \eqref{eq:nonlinear-rep-learning}: 
\begin{align}
\bm{x}_i &= f(m^{-1}(m(\bz_i))) + \bm{\varepsilon}_i,
\quad\text{for any invertible $m: \mathcal{Z} \to \mathcal{Z}$.}
\label{eq:m-transform}
\end{align}
Thus, without restrictions on $\fcns$ or $\zprs$, the model is trivially nonidentifiable.

To clarify this, let $m_{\#}\zpr$ denote the push-forward measure of $\zpr$ by $m$, i.e. the distribution on $\bz_i$ induced by $m$. Then if both  $f\circ m^{-1} \in \fcns$ and  $m_{\#}\zpr\in\zprs$ for some non-identity $m$, then the model is not identifiable. Following the terminology of \citet{xi2023indeterminacy}, we refer to such transformations $m$ that satisfy both (i) and (ii) as indeterminancy transformations; the set of indeterminancy transformations is the indeterminancy set.
\vspace{0.5em}
\begin{definition}\label{defn:indetermin}
Given $(f, \zpr) \in \funcs \times \zprs$, $m:\mathcal{Z}\to\mathcal{Z}$ is an \emph{indeterminancy transformation} if $f\circ m^{-1} \in \funcs$ and $m_{\#}\zpr \in \zprs$. The \emph{indeterminancy set}, $\mathcal{A}(\funcs, \zprs)$, is the set of all indeterminancy transformations of $(\funcs, \zprs)$. 
\end{definition}

It is clear that the identity transformation $m=\mathrm{id}$ is always in the indeterminacy set $\mathcal{A}(\funcs, \zprs)$.
When $\mathcal{A}(\funcs, \zprs)$ contains additional transformations, then the model parameters are not identifiable. To resolve identifiability issues, we can again further restrict the function class $\fcns$, the prior distributions $\zprs$, or both.  We state this more precisely in \Cref{defn:ident}; for a formal measure-theoretic treatment, see \citet{xi2023indeterminacy}.

\vspace{0.5em}
\begin{definition}
\label{defn:ident}
Let $\zprs$ be a family of probability distributions on $\mathcal{Z}$ and $\funcs$ be a family of functions $f:\mathcal{Z}\rightarrow \mathcal{X}$.  
\begin{enumerate}
\item For the model parameterized by $(f, \zpr)\in \funcs \times \zprs$, we say that $f$ is \emph{identifiable} (from $f_\#\zpr$) \emph{up to} $\mathcal{A}(\funcs, \zprs)$. 
\item The model parameters $(f, \zpr)$ are \emph{strongly identifiable} (from $f_\#\zpr$) if $\mathcal{A}(\funcs, \zprs) = \{\mathrm{id}\}$; i.e. the identity function on $\mathcal{Z}$ is the only indeterminancy transformation. 
\end{enumerate}
\end{definition}

In general, obtaining sufficient conditions for identifying $(f, \zpr)$ is easy; we can simply place many restrictions on $\fcns$ or $\zprs$. In the linear case, for example, we can constrain the loadings matrix to be upper triangular. The difficulty is finding sufficient conditions that identify the model while remaining as flexible as possible.

Furthermore, we often relax the requirement of strong identifiability. For example, we may not be interested in the ordering of the latent variables and so the indeterminancy set can contain permutations of the labels (``label switching''). Other common indeterminacies include scaling (diagonal matrices), translations, and orthogonal transformations.
A common goal in machine learning is ``disentanglement'' \citep{bengio2013deep}, which roughly means to separate distinct latent factors of variation.
\Cref{eg:dis} below helps formalize the relationship between identifiability and ``disentangling'' latent factors of variation.
\vspace{0.5em}
\begin{definition}[Disentanglement] \label{eg:dis}
Consider a learned representation $\widehat{\bz} = h(\bm{x})$ where $h:\mathbb{R}^\xdim\to\mathbb{R}^\zdim$ is an estimate of $f^{-1}$. Then $\widehat{\bm{z}}$ is said to be ``disentangled'' with respect to the ground-truth factors $\bz$  if $\widehat{\bz}$ is an element-wise transform of $\bz$, up to permutation:
\begin{align}
(\widehat{z}_{\sigma(1)},\dots, \widehat{z}_{\sigma{(K)}}) = (g_1(z_1),\dots, g_K(z_K)),
\end{align}
where $\sigma$ is a permutation and $g_k$ are invertible scalar functions \citep{ahuja2022weakly,yao2024unifying}. This is equivalent to the indeterminancy set $\mathcal{A}(\mathcal{F}, \mathcal{P})$ containing only element-wise transformations and permutations.
\end{definition}

\bigskip
\begin{remark}
In the earlier ML literature, ``disentanglement'' referred to finding independent factors of variation \citep[e.g.][]{bengio2013deep}. Unfortunately, independence is insufficient for identification, a fact that is well-known in the ICA literature \citep{hyvarinen1999nonlinear}.
This was emphasized in the disentanglement context by \citet{locatello2019challenging}. 
These results rely on the classical Darmois construction, which uses an orthogonal transformation to keep the prior $\zpr$ invariant while rotating the mixing function $f$.
Consequently, there was a shift towards identification-based definitions of disentanglement. Eventually, dependent factors of variation began to be considered alongside the development of CRL. \end{remark}

\subsubsection{Identifiability assumptions on the mixing function}\label{sec:mixing-id}

We now discuss approaches which place assumptions on the mixing function class $\fcns$ to identify the model.  

\noindent
\parhead{Anchor features.} In the linear case, we saw how anchor features can eliminate rotation indeterminancies. \citet{moran2022identifiable} extend this assumption to the nonlinear case  by introducing a masking variable $\bm{w}_j \in \{0,1\}^K$ for each feature $j\in [D]$: 
\begin{align}
x_{ij} = f(\bm{w}_j \odot \bz_i)  + \varepsilon_{ij}, \label{eq:sparse-vae-moran}
\end{align}
where $\odot$ denotes the element-wise product. The masking variable $\bm{W} = [\bm{w}_1^T, \cdots, \bm{w}_D^T]$ is analogous to the sparse loadings matrix in the linear setting. With the anchor assumption, and additional assumptions on the factor covariance and  $f$, \citet{moran2022identifiable} prove that the latent factors are identifiable up to element-wise transformations and permutations. 
See also \citet{xu2023identifiable}, who obtain identifiability with an anchor feature assumption, additionally assuming  $f$ is a generalized additive model and independent latent variables. Another line of work uses anchor features (equivalent to so-called \emph{pure children}) to identify latent hierarchical models via rank constraints~\citep{huang2022latent,kong2023identification}.

\noindent
\parhead{Subset condition.}
Around the same time, \citet{kivva2021learning} proposed the ``subset condition'' for identifying discrete latent variables in general nonlinear models, and showed that it is nearly necessary for identification. The subset condition requires that the observed children of $z_{ij}$ are not a subset of the observed children of $z_{ik}$ for any $k\ne j$, which can be interpreted as a sparsity condition on $f$. In particular, anchor features are a special case of the subset condition. This was later extended to continuous features in \citet{kivva2022identifiability}, which established the identifiability of a universal class of functions (specifically, piecewise linear $f$), addressing an open problem from \cite{wang2021posterior}.

There have also been recent works on nonlinear ICA which exploit sparsity in $f$ for identifiability \citep{zheng2022identifiability,zheng2023generalizing}, although ICA models do not consider dependence between the latent factors. These works assume a ``structural sparsity'' constraint on $f$, which is equivalent to the subset condition of \citet{kivva2021learning}.  Also in the independent factors setting, \citet{markham2020measurement,markham2024ncfa} prove identifiability if a certain dependence graph over $\bx_{i}$ admits a unique minimum edge clique cover, for which the anchor feature assumption is sufficient but not necessary.

\noindent
\parhead{Independent mechanisms.}  The principle of \emph{independent mechanisms} (\Cref{sec:causal:ivn}) is  also related to sparsity and classical assumptions from PCA. \citet{gresele2021independent} operationalize this principle by assuming that the mixing function $f$ has a Jacobian with orthogonal columns, proving that this assumption eliminates certain classes of mixing functions from the set of indeterminacy transformations, $\mathcal{A}(\funcs, \zprs)$ when $\zdim = \xdim$; \citet{ghosh2023independent} extend this result to $\zdim \ll \xdim$.  In the linear case, this assumption corresponds to orthogonality conditions on the loadings matrix, as in PCA.

\subsubsection{Identifiability assumptions on the latent factors}

An alternative strategy to identify the model is to restrict the factor distribution $\zpr$.

\noindent
\parhead{Temporal dependence.}
\citet{hyvarinen2016unsupervised} popularized using temporal structure to identify latent factors; specifically, they assume non-stationarity of the latent factors over time to obtain identifiability for nonlinear ICA.   For stationary latent distributions, \citet{halva2021disentangling} prove that $f$ is identifiable up to element-wise transform and permutation. More generally, \citet{ahuja2021properties} consider how temporal dependencies restrict the indeterminancy set. \citet{lachapelle2022disentanglement} assume sparsity in the latent transition graph to prove element-wise identifiability up to permutations.  \citet{lippe2022citris} assume access to known latent interventions at each time step to prove element-wise identifiability of $f$; \citet{lippe2023biscuit} allow for unknown binary interventions.

\noindent
\parhead{Auxiliary information.}
Temporal structure can be viewed as a type of  auxiliary data. General auxiliary data approaches have been considered by \citet{hyvarinen2019nonlinear, khemakhem2020variational}, who assume that in addition to $\bm{x}_i$, another variable $\bm{u}_i$ is observed, and $p(\bz_i|\bm{u}_i)$ is an exponential family. This restriction identifies $\zpr$ up to a linear transformation, which can be improved to element-wise transformations when the exponential family has sufficient statistic dimension greater than two. \citet{hyvarinen2019nonlinear} use self-supervised learning, while \citet{khemakhem2020variational} take a generative model approach.

\noindent
\parhead{Mixture priors.}
In addition to addressing identifiability under universal approximation,
\citet{kivva2022identifiability} also eliminate the need for auxiliary data; instead, they show how mixture priors enable identification.
When $\zpr$ is a Gaussian mixture and the clusters are distinct, this prevents arbitrary rotations of $\zpr$, identifying the factors up to element-wise linear transformations (and permutations). \citet{kivva2022identifiability} also prove identifiability of $f$ up to affine transformations, assuming $f$ is piecewise affine and injective.

\noindent
\parhead{Invariance.}
Identifiability can also be achieved by exploiting the existence of invariant structure in the latents under data augmentations or multiple data views. \citet{von2021self} assume access to augmentations of the observed data $(\bm{x}, \widetilde{\bm{x}})$ where a subset of the underlying factors for $\bm{x}$ and $\widetilde{\bm{x}}$ are the same (i.e., some elements of $\bz, \widetilde{\bz}$ are equal, often enforced via contrastive learning); this is extended to multiple augmentations in \citet{eastwood2023self}. \citet{yao2024multiview} assume access to multiple views of the data; similarly to data augmentations, the latent variables can also be identified in this case.

\subsection{Causal identifiability through interventional data}\label{sec:intervention-id}

So far, we have considered the statistical identifiability of $(f, \zpr)$. 
We now move onto causal identifiability of $\latgr$ and/or $\latsem$ through the use of richer types of data such as interventions.
As discussed in \Cref{sec:crl-identifiability}, we generally cannot infer $\latgr$ from observational data alone.
In fact, even when $\bm{x}_i$ is linear in $\bz_i$, \citet{squires2023linear} show that it is impossible to recover $\latgr$
without at least one intervention on each latent factor.

This more difficult regime with interventions is called \emph{interventional causal representation learning}.
We now discuss identifiability results in interventional CRL in both the linear mixing case $\bm{x}_i=\bm{B}\bz_i$, and  the nonlinear mixing case $\bm{x}_i=f(\bz_i)$. 
Although we emphasize interventions here, identification through more general types of distribution shifts is an active area of research (Section~\ref{sec:conc}).

\noindent
\parhead{Linear mixing.}  \citet{squires2023linear} consider the linear mixing case where $\funcs$ is the set of full rank linear transformations, parametrized by matrices $\bm{B}\in\R^{\xdim \times \zdim}$:
\begin{align}
\bm{x}_i = \bm{B} \bm{z_i} + \bm{\varepsilon}_i, \quad i = 1,\dots, N, \label{eq:squires-linear-model}
\end{align}
with $\bz_i$ satisfying the linear SEM \eqref{eq:linear-sem-crl} and where $\bm{\varepsilon}_i$ has independent components.  (Note that $\bz_i$ does not need to be Gaussian). 
\citet{squires2023linear} assume the observed data is $\{\bm{x}_{1:N}, \bm{x}_{1:N_e}^{(e)}\}_{e\in\mathcal{E}}$ where $\bm{x}_i$ is drawn from the observational distribution and $\bm{x}_i^{(e)}$ is drawn from an interventional distribution, and prove that the latent causal graph $\latgr$ is identifiable when $\mathcal{E}$ are single-node perfect interventions acting on each of the latent variables (see \Cref{def:single-node}).

The key idea behind the identification strategy of \citet{squires2023linear} is that the source nodes of the graph $\gr$ (i.e. the latent variables without any parents) leave observable signatures in the data $\{\bm{x}, \bm{x}^{(e)}\}$. Specifically, if $\bm{x}^{(e)}$ corresponds to a source node intervention, then the difference in the precision matrices of $\bm{x}$ and $\bm{x}^{(e)}$ will be rank-one.  This is because if a source node is intervened upon, all the downstream dependencies will be the same as the observational data; only the source node distribution is modified. This strategy can be applied iteratively to identify the partial ordering (i.e. topological sort) of the latent causal graph. When the interventions are perfect, this partial ordering provides a constraint which identifies the causal graph of $\bm{z}_i$.

\noindent
\parhead{Nonlinear mixing.} \citet{buchholz2023learning} generalize the results of \citet{squires2023linear} to the nonlinear setting; specifically, they consider nonlinear mixing functions $f$ which are injective and differentiable. For the latent factors $\bz_i$, \citet{buchholz2023learning} also consider a linear SEM  \eqref{eq:linear-sem-crl}, but restrict $\bz_i$ to be  Gaussian.  \citet{buchholz2023learning} prove element-wise identifiability of the latent $\bz_i$, as well as the causal graph of $\bz_i$, again with the assumption of single-node perfect interventions on all the latent variables. The key idea is the same: The difference between the observational and source node precision matrices is still rank-one; this geometric structure allows for $f$ to be identified up to linear transformations, at which point the results of \citet{squires2023linear} can be applied.

\citet{zhang2023identifiability} also consider nonlinear mixing functions and a linear SEM on $\bz_i$, but allow for soft interventions instead of perfect interventions on $\bz_i$. As soft interventions do not remove connections to parent variables, the causal graph can only be identified up to ancestors \citep[formally, the transitive closure;][]{tian2001causal}. Consequently, \citet{zhang2023identifiability} need an additional sparsity assumption on the graph for identifiability. 

\noindent
\parhead{Beyond linear SEM.}
So far we have assumed a linear SEM on $\bz_i$, although this is not really necessary. 
\cite{jiang2023learning} study a general setting with no parametric assumptions and no functional constraints, and provide a minimal set of nonparametric assumptions under which 
the latent causal graph $\latgr$ can be identified up to so-called ``isolated edges''. These assumptions are minimal in the sense that if any single assumption is violated, then counterexamples to identifiability can be constructed.

One consequence of avoiding functional constraints is that statistical identifiability becomes challenging. To overcome this, \cite{varici2024general,von2023nonparametric} require at least two perfect interventions per latent node as well as  ``interventional discrepancy'', a requirement that the interventional distributions differ almost everywhere. \citet{von2023nonparametric} require knowing which two interventions operate on the same latent variable; \citet{varici2024general} do not need to know which interventions are paired.   The intuition is: for datasets arising from the same intervention, their corresponding latents must have the same distribution, except for the intervened latent dimension. This constrains the mapping from the data to the latent space, enabling identification of the latents. Then, the latent causal graph can be obtained by comparing interventional and observational data.

A different line of work considers an ``independent support'' condition to identify latent variables \citep{wang2024desiderata,ahuja2023interventional}. These papers make an overlap assumption in the latent space; that is, there is non-zero probability on all combinations of latent factors. To provide intuition for this assumption, it is helpful to take an experimental design perspective; we need a design table where all cells (combinations of latent factor levels) have observations. The overlap assumption means that the support of the latent factors is independent, or rectangular. This again constrains the mapping from the data to the latent space, enabling element-wise identification of $\bm{z}$.

\section{Estimation in causal representation learning}
\label{sec:practical}

In this section, we discuss current methodological approaches to estimation in order to provide an overview of how CRL can be implemented in practice. 

\noindent
\parhead{Variational autoencoders.}
A popular approach to fit latent nonlinear representations is the variational autoencoder \citep[VAE,][]{kingma2013auto,rezende2014vae}. Consider the deep generative model (DGM):
\begin{align}
\bm{z}_i &\sim N(\bm{0}, \bm{I}) \notag \\
\bm{x}_i &= f_{\theta}(\bm{z}_i) + \varepsilon_i, \quad \varepsilon_i \sim N(\bm{0}, \bm{I}). \label{eq:sec-6-dgm}
\end{align}
A VAE approximates the latent posterior distribution  $p(\bm{z}_i | \bm{x}_i)$ with the variational density:
\begin{align*}
q_{\varphi}(\bm{z}_i|\bm{x}_i) \sim \mathcal{N}_K(\mu_{\varphi}(\bm{x}_i), \sigma_{\varphi}^2(\bm{x}_i)),
\end{align*}
where $\mu_{\varphi}, \sigma_{\varphi}^2: \R^D \to \R^K$ are typically neural networks. Although the outputs of these neural networks are low-dimensional, the networks themselves can be arbitrarily complex and consist of any number of hidden neurons and layers.

As in standard variational inference, the evidence lower bound (ELBO) on the marginal log-likelihood is optimized with respect to the parameters $\{\theta, \varphi\}$:
\begin{align*}
\log p(\bm{X}) \geq \sum_{i=1}^N \E_{q_{\varphi}(\bm{z}_i|\bm{x}_i)}[\log p_{\theta}(\bm{x}_i|\bm{z}_i)] - D_{\mathrm{KL}}(q_{\varphi}(\bm{z}_i|\bm{x}_i) || p(\bm{z}_i)).
\end{align*}
The first term of the ELBO involves an intractable expectation; this expectation is typically approximated with $L$ Monte Carlo samples of the latent variables:
\begin{align*}
    \bm{z}_i^{(l)} = \mu_{\varphi}(\bm{x}_i) + \sigma_{\varphi}(\bm{x}_i)\odot \bm{\varepsilon}_{i}^{(l)}, \quad \bm{\varepsilon}_i^{(l)} \sim \mathcal{N}_K(\bm{0}, \bm{I}), \quad l = 1,\dots, L.
\end{align*}
Then, the approximate ELBO is optimized over $\{\theta, \varphi\}$ with stochastic gradient ascent.

\noindent
\parhead{Identifiable representation learning.}
The DGM in \eqref{eq:sec-6-dgm} is not identifiable. As discussed in \Cref{sec:mixing-id}, \citet{moran2022identifiable} introduce a masking variable $\bm{W}$ \eqref{eq:sparse-vae-moran}; when an anchor assumption holds, the model is identifiable. In practice, a sparsity-inducing prior $p(\bm{W})$ can be placed on the masks.  The resultant ELBO is:
\begin{align*}
\mathcal{L}(\theta, \varphi, \bm{W},\bm{\rho}) &= \sum_{i=1}^N \E_{q_{\varphi}(\bm{z}_i|\bm{x}_i)}[\log p_{\theta}(\bm{x}_i|\bm{z}_i, \bm{W})] - D_{\mathrm{KL}}(q_{\varphi}(\bm{z}_i|\bm{x}_i) || p(\bm{z}_i)) +  \log p(\bm{W}).
 \end{align*}
When the prior $p(\bm{W})$ is hierarchical \citep[e.g. the Spike-and-Slab Lasso of ][]{rockova2018spike}, the additional prior parameters may be estimated using the EM algorithm \citep{moran2022identifiable}.
This sparse VAE can be viewed as a nonlinear analogue to factor analysis as  $\widehat{\bm{W}}$ can be interpreted similarly to a loadings matrix: for each column of $\widehat{\bm{W}}$, the non-zero features can be considered a cluster, as they all depend on the same factor dimension.

\noindent
\parhead{Interventional causal representation learning.}
\citet{zhang2023identifiability} propose a method for interventional CRL,
the CausalDiscrepancy VAE. The latent $\bm{z}_i$ follow a linear SEM, and the observed data is a nonlinear function of the latent variables:
\begin{align}
    \bm{z}_i &= (I
    -\bm{A})^{-1}\bm{\nu}_i, \quad \bm{\nu}_i \sim \mathcal{N}_K(0, I) \\
    \bm{x}_i &= f_{\theta}(\bm{z}_i) + \bm{\varepsilon}_i,\quad \bm{\varepsilon}_i\sim \mathcal{N}_D(0, I),  
\end{align}
where $\bm{A}\in\R^{K\times K}$ captures the (unknown) causal graph among the $\bm{z}_i$.  The CausalDiscrepancy VAE estimates the exogenous noise $\bm{\nu}_i$ distribution with the variational approximation $q(\bm{\nu}_i|\bm{x}_i) \sim \mathcal{N}_K(\mu_{\varphi}(\bm{x}_i), \sigma_{\varphi}^2(\bm{x}_i))$, from which $\bm{z}_i$ is obtained as $\widehat{\bm{z}}_i = (1-\widehat{\bm{A}})^{-1}\widehat{\bm{\nu}}_i$. 

The CausalDiscrepancy VAE also learns  a map from the intervention label to the latent target and intervention shift: $T_{\phi}:[|\mathcal{E}|]\to \{[K], \R\}$. This $T_{\phi}$ allows ``virtual'' interventional data to be generated and compared to actual interventional data. The CausalDiscrepancy VAE combines the usual VAE objective with (i) a maximum mean discrepancy \citep[MMD, ][]{gretton2012kernel} term that encourages data from the virtual interventional distribution $\widetilde{\bm{X}}^{(e)}\sim {P}_{\widehat{\theta}, \widehat{\phi}}^{(e)}$ to be close to the actual interventional data $\bm{X}^{(e)}$, and (ii) a sparsity penalty on the causal graph $\latsem$:
\begin{align}
\mathcal{L}(\theta, \varphi, \phi, A) &= \sum_{i=1}^N \E_{q_{\varphi}(\bm{\nu}_i|\bm{x}_i)}[\log p_{\theta}(\bm{x}_i|\bm{z}_i)] - D_{\mathrm{KL}}(q_{\varphi}(\bm{\nu}_i|\bm{x}_i) || p(\bm{\nu}_i)) \\
&\quad - \mathrm{MMD}(\widetilde{\bm{X}}^{(e)}||\bm{X}^{(e)}) - \lambda \lVert A\rVert_1.
\end{align}

\citet{zhang2023identifiability} apply the CausalDiscrepancy VAE to Peturb-Seq data \citep{norman2019exploring}. In addition to an observational dataset, there are interventional datasets, each corresponding to a single-gene CRISPR activation. The function $T_{\phi}$ maps the intervention label (gene label) to a latent dimension; the latent dimensions can then be interpreted by inspecting which gene interventions are mapped to the same latent variable.

\noindent
\parhead{Score-based methods.} \citet{varici2024general} also propose a method for CRL from interventional data. Unlike the previous methods, \citet{varici2024general} do not use a VAE-based framework and instead rely on (Stein) score functions of the data. Specifically, for two interventions $e$ and $\widetilde{e}$, the difference in latent score functions is:
\begin{align}
\nabla_{\widehat{\bm{z}}} \log p^{(e)}(\widehat{\bm{z}}) - \nabla_{\widehat{\bm{z}}}\log p^{(\widetilde{e})}(\widehat{\bm{z}}) = \bm{J}_{h^{-1}}(\widehat{\bm{z}})[\nabla_{\bm{x}} \log p^{(e)}(\bm{x}) - \nabla_{\bm{x}} \log p^{(\widetilde{e})}(\bm{x})],
\end{align}
for any function $h$ with $\widehat{\bm{z}} = h(\bm{x})$. When $e$ and $\widetilde{e}$ intervene on the same latent dimension, the above difference is non-zero only for that dimension. This yields a constrained optimization problem for learning $h$, the solution of which identifies $\bm{z}$. The causal graph of $\bm{z}$ can then be learned by comparing score functions of observational and interventional data.

\section{Conclusion and open questions}
\label{sec:conc}

CRL is an emerging discipline with roots in machine learning, statistics, and causality, and draws inspiration from a variety of domains to address important practical challenges in the use of generative models in scientific applications where interpretability and causality are key. From a statistical perspective, CRL offers a new twist on classical factor analysis through its principled use of causal semantics (graphs, interventions, etc.).

Although the ideas behind CRL have evolved slowly over many decades, it is still a developing discipline with numerous open questions. We conclude with a brief summary of some directions for future inquiry.

\noindent
\parhead{Beyond interventions.}So far we have focused on using interventions for identification. 
While this is indeed practical in a variety of settings, there is significant interest in generalizing beyond interventions.
An emerging paradigm is to use multiple \emph{environments} (i.e. non-iid data arising from data augmentation, distribution shifts, different locations, times, contexts, etc.) to identify latent structure and/or improve generalization \citep{peters2016causal,mooij2020joint,huang2020causal,perry2022causal}.
Indeed, experimental interventions are a special type of environment.
It is an intriguing problem to understand how general environments can enable identifiability when strict interventions are not available; see e.g. \citet{varici2024general,chen2024identifying}.

\noindent
\parhead{Implementation and evaluation.}
Early work on CRL has mostly focused on the identification problem; however, just as important is the problem of estimation and practical algorithms. Despite substantial advances in practical aspects of training generative models, these advances have not directly translated over to the setting of CRL. Recent work (Section~\ref{sec:practical}) has begun to address these challenges, but no general framework has emerged. Another important problem is evaluating CRL models: Since we can only identify parameters up to indeterminacy transformations, how can we assess the quality of estimators in a principled way? One approach is to use downstream tasks such as out-of-distribution generalization \citep{lu2021invariant} 
and extrapolation \citep{saengkyongam2024identifying}.

\noindent

\parhead{Optimization and nonconvexity.}
A significant challenge in implementing and evaluating CRL models is that the optimization landscape is highly nonconvex and sensitive to modeling assumptions. This is true even in the linear case, as is well-known for factor models, and these problems are only exacerbated with neural networks. Moreover, in practice, classical nonparametric estimators  are replaced with variational approximations that are optimized through stochastic gradient descent. The behaviour of these approximations is an outstanding open problem of significant interest, even for simple one-dimensional generative models \citep[see e.g.][]{kwon2024minimax}. On a positive note, identifiable models often have better-behaved loss surfaces (e.g. well-defined minima); investigating the optimization landscape in identifiable CRL models is an interesting question for further research.

\noindent
\parhead{Statistical theory.}
There are significant gaps in our statistical understanding of CRL: Finite-sample rates, uncertainty quantification, and misspecification all remain underexplored. 
Some recent exceptions in this direction include \citet{acarturk2024sample,buchholz2024robustness}, although many problems remain:
Since the models are nonparametric, the usual curse of dimensionality applies, but can causal assumptions help mitigate the curse of dimensionality? 
Alternatively, if we aim only to estimate the graph $\latgr$ and its parameters $\latsem$, are conventional rates achievable?  
What about robustness and estimation under misspecification?
These are just some of the intriguing statistical questions that demand resolution.

\noindent
\parhead{Generalization and transfer.}
Transferability and out-of-distribution (OOD) generalization are key motivations for learning causal representations in practice, which are related to recent work on causal invariance \citep{buhlmann2020invariance} and invariant causal prediction \citep{peters2016causal}. It remains an important statistical problem to formalize these connections and to understand how and when causal representations lead to genuine improvements in OOD transfer and generalization. 
Numerous studies point to challenges when using non-causal models \citep{schott2021visual,montero2021role}, and recent work suggests causal models as a potential solution \citep{kong2022partial,richens2024robust}.

\noindent
\parhead{Large language models.}
The development of large language models (LLMs) and more generally, foundation models, has occurred in parallel with CRL. This suggests the tantalizing possibility of developing pre-trained foundation models with causal interpretability. This comes with numerous practical challenges; e.g. connecting the objectives of next-token prediction with causal objectives; principled use of diverse, multi-modal, internet-scale datasets; extracting interpretable concepts; etc. On the latter, an emerging literature explores how to identify interpretable concepts in foundation models  \citep{leemann2023post,rajendran2024learning}. This connects to a broader literature on how concepts are embedded in the latent spaces of foundation models via so-called \emph{linear representations} \citep[e.g.][]{mikolov2013linguistic,szegedy2013intriguing}. 
An alternative line of work extracts interpretable concepts post-hoc from pre-trained LLMs using sparse dictionary learning \citep{bricken2023monosemanticity,templeton2024scaling}. 
These sparse dictionary learning approaches assume that each token is the linear combination of a small set of concept vectors, selected from a overcomplete concept dictionary. That is, the concept space has a much higher dimension than the data, in contrast to most CRL methods. Investigating these higher-dimensional latent space settings in the context of CRL is an important area for future research.

\section{Funding}\label{funding-statement}

B.A. was supported by NSF IIS-2453378, NSF IIS-1956330, and NIH R01GM140467.

\bibliography{bib/ref}

\end{document}